\documentclass[letterpaper]{article} % DO NOT CHANGE THIS
\usepackage{aaai2026}  % DO NOT CHANGE THIS
\usepackage{times}  % DO NOT CHANGE THIS
\usepackage{helvet}  % DO NOT CHANGE THIS
\usepackage{courier}  % DO NOT CHANGE THIS
\usepackage[hyphens]{url}  % DO NOT CHANGE THIS
\usepackage{graphicx} % DO NOT CHANGE THIS
\usepackage{natbib}  % DO NOT CHANGE THIS AND DO NOT ADD ANY OPTIONS TO IT
\usepackage{caption} % DO NOT CHANGE THIS AND DO NOT ADD ANY OPTIONS TO IT
\usepackage{algorithm}
\usepackage{algorithmic}
\usepackage{amsfonts}
\usepackage{graphicx}
\usepackage{subcaption}
\usepackage{multirow}
\usepackage{booktabs}
\usepackage{amsmath}
\usepackage{minted}
\usepackage{newfloat}
\usepackage{listings}
\DeclareCaptionStyle{ruled}{labelfont=normalfont,labelsep=colon,strut=off} % DO NOT CHANGE THIS
\floatstyle{ruled}
\newfloat{listing}{tb}{lst}{}
\floatname{listing}{Listing}
\title{Language Drift in Multilingual Retrieval-Augmented Generation: Characterization and Decoding-Time Mitigation}
\author{
    Bo Li\textsuperscript{\rm 1, 2},
    Zhenghua Xu\textsuperscript{\rm 1},
    Rui Xie\textsuperscript{\rm 2}\thanks{Corresponding Author.}
}
\affiliations{
    \textsuperscript{\rm 1}State Key Laboratory of Intelligent Power Distribution Equipment and System, \\School of Health Sciences and Biomedical Engineering, Hebei University of Technology, China\\
    \textsuperscript{\rm 2}National Engineering Research Center for Software Engineering, Peking University, China\\
    \texttt{deepblue.lb@gmail.com, ruixie@pku.edu.cn}
}
\usepackage{bibentry}
\begin{document}

\maketitle

\begin{abstract}
Multilingual Retrieval-Augmented Generation (RAG) enables large language models (LLMs) to perform knowledge-intensive tasks in multilingual settings by leveraging retrieved documents as external evidence. However, when the retrieved evidence differs in language from the user query and in-context exemplars, the model often exhibits language drift by generating responses in an unintended language. This phenomenon is especially pronounced during reasoning-intensive decoding, such as Chain-of-Thought (CoT) generation, where intermediate steps introduce further language instability. In this paper, we systematically study output language drift in multilingual RAG across multiple datasets, languages, and LLM backbones. Our controlled experiments reveal that the drift results not from comprehension failure but from decoder-level collapse, where dominant token distributions and high-frequency English patterns dominate the intended generation language. We further observe that English serves as a semantic attractor under cross-lingual conditions, emerging as both the strongest interference source and the most frequent fallback language. 

To mitigate this, we propose Soft Constrained Decoding (SCD), a lightweight, training-free decoding strategy that gently steers generation toward the target language by penalizing non-target-language tokens. SCD is model-agnostic and can be applied to any generation algorithm without modifying the architecture or requiring additional data. Experiments across three multilingual datasets and multiple typologically diverse languages show that SCD consistently improves language alignment and task performance, providing an effective and generalizable solution in multilingual RAG.
%Multilingual Retrieval-Augmented Generation (RAG) enables large language models to answer questions by reasoning over retrieved evidence. However, when the retrieved documents are in a different language than the query and examples, models often exhibit output language drift, generating responses in an unintended language. In this paper, we systematically investigate the phenomenon of output language drift, where large language models generate responses in a language different from the user query and in-context examples. Through controlled experiments across multiple languages, models, and datasets, we reveal that this drift intensifies during Chain-of-Thought (CoT) generation and often defaults to English, regardless of the context language. Our analysis shows that this behavior is not a comprehension failure, but a decoding-stage collapse caused by dominant token distributions. To address this, we introduce Soft Constrained Decoding (SCD), a model-agnostic, decoding-time method that softly penalizes non-target-language tokens. SCD improves both language consistency and task performance, offering a lightweight yet effective solution for multilingual RAG.
\end{abstract}

\begin{links}
    \link{Code and Dataset}{https://github.com/pkuserc/SCD}
\end{links}
% Uncomment the following to link to your code, datasets, an extended version or similar.
% You must keep this block between (not within) the abstract and the main body of the paper.
% \begin{links}
%     \link{Code}{https://aaai.org/example/code}
%     \link{Datasets}{https://aaai.org/example/datasets}
%     \link{Extended version}{https://aaai.org/example/extended-version}
% \end{links}
\section{Introduction}

Recent advances in Retrieval-Augmented Generation (RAG) have significantly enhanced large language models’ ability to generate factually grounded answers in open-domain question answering~\cite{Xu2024UnsupervisedIR,Luo2024LandmarkEA,Fang2024EnhancingNR,shi2024generate}. However, in multilingual settings, most existing studies have focused on improving cross-lingual retrieval performance and contextual alignment~\cite{Luo2020VECOVA,Park2025InvestigatingLP,Chirkova2024RetrievalaugmentedGI,Wu2024NotAL,Ranaldi2025MultilingualRG,Ranaldi2025ImprovingMR,Liu2025XRAGCR,Blandon2025MEMERAGAM}, while overlooking a critical issue: the mismatch between the input and output languages. 

\begin{figure}[t]
	\centering
	\includegraphics[width=0.99\linewidth]{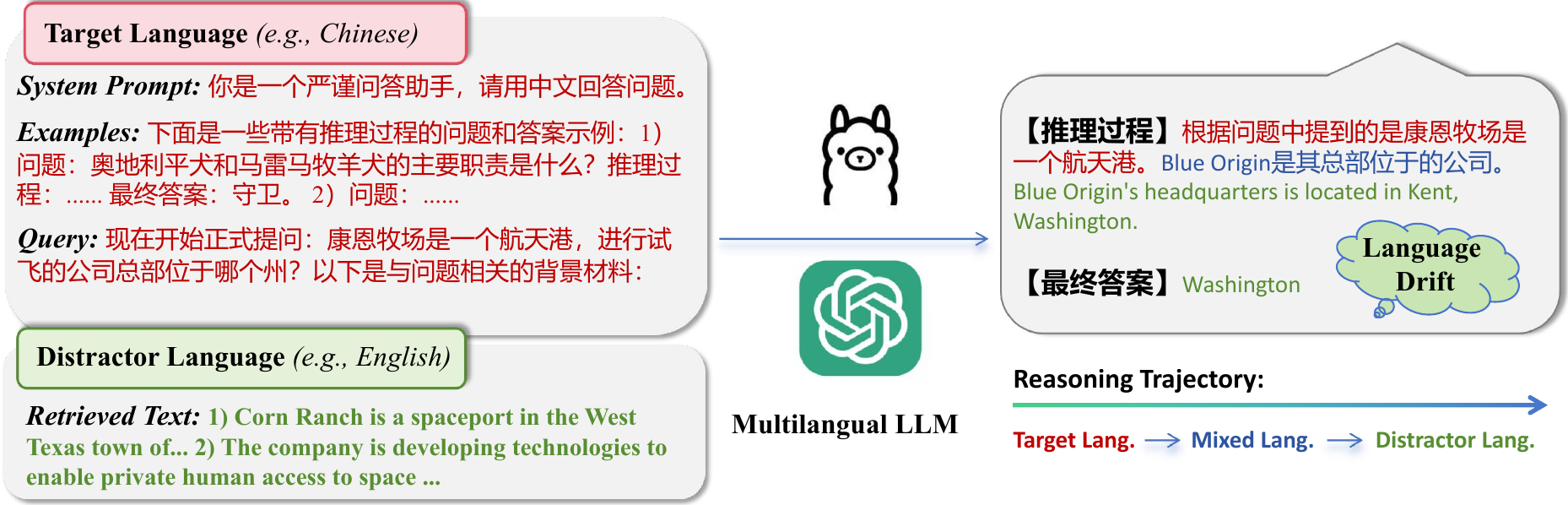}
	\caption{ Illustration of language drift in multilingual RAG. The user query and in-context examples are provided in the target language (e.g., Chinese), while the retrieved context is written in a non-target language (e.g., English). During reasoning, the model mixes languages and ultimately outputs the final answer in a non-target language.}
	\label{pic:intro}
\end{figure}

In multilingual RAG settings, the query, instructions, and in-context exemplars are typically written in the target language, aiming to elicit responses in that language. However, due to the predominance of English in open-domain corpora~\cite{Xu2024ASO,Zeng2024EnglishLH,Resnik2003TheWA}, the retrieved context is often in English, even when the query is in another language. This creates a mixed-lingual input scenario where only the retrieved context differs in language. Nevertheless, models frequently generate responses in the language of the retrieved content rather than in the intended target language~\cite{Park2025InvestigatingLP,Liu2025XRAGCR}. We refer to this phenomenon as \textit{output language drift}, which poses practical challenges for multilingual applications yet remains underexplored. Our empirical study reveals that such cross-lingual conditions negatively affect both task performance and output language consistency. Notably, English serves as the strongest interference source, significantly degrading output quality in non-English settings, while also serving as the most robust target language when subjected to interference. This issue becomes more severe under few-shot prompting and Chain-of-Thought (CoT) reasoning~\cite{Shi2022LanguageMA,Yu2025CrossLingualCA}. %where the model, despite receiving instructions and exemplars in the target language, still integrates information from non-target evidence during generation. 

Interestingly, we observe that language drift does not necessarily follow the language of the retrieved context. Instead, models frequently default to English during generation, even when the context passages are in Arabic, Russian, or other non-English languages. This indicates that English plays a dominant role beyond being a common training language: it functions as a semantic attractor in multilingual generation. Our analysis indicates that, under cross-lingual ambiguity, LLMs tend to prefer English over the context language. This fallback tendency further verifies the dominant role of English as the default trajectory in multilingual decoding.

%Crucially, this language drift does not always align with the context language. Instead, we find that models often default back to English during generation, even when context passages are in languages such as Arabic or Russian. This suggests that English serves as more than just a frequent context language; it acts as a semantic attractor in multilingual generation. Our analyses show that under language conflict or ambiguity, LLMs are more likely to revert to English than to mimic the language of the input context. This default behavior highlights the dominant role of English as a fallback generation path.

%To investigate whether this behavior stems from comprehension failure or decoding bias, we design settings where the query, prompt, and exemplars are consistently in the target language, with only the retrieved evidence presented in a different language. 

To better understand whether this fallback behavior results from misunderstanding or from generative biases, we conduct human evaluation and reference translation. Interestingly, many of the outputs that drift to the non-target language are still semantically faithful, indicating that the model has accurately understood both the task and the retrieved context. By analyzing intermediate reasoning steps (i.e., CoT traces), we find that the language inconsistency often emerges mid-generation, even when earlier steps remain in the target language. This indicates that the failure stems not from semantic comprehension, but from generation biases favoring frequent English tokens. As a result, the model produces outputs that are structurally fluent but linguistically inconsistent, reflecting a form of language collapse driven by token-level priors rather than task misunderstanding.

These findings motivate the need for lightweight decoding-time strategies that maintain output language consistency without compromising reasoning performance. To this end, we introduce Soft Constrained Decoding (SCD), a token-level control mechanism that assigns soft penalties to non-target-language tokens, thereby encouraging target-language generation while preserving fluency. In contrast to hard vocabulary filtering, SCD is a flexible, model-agnostic mechanism compatible with standard decoding algorithms. Extensive experiments across diverse datasets, model backbones, target languages, and context languages demonstrate that SCD improves both output language alignment and answer quality, providing a practical solution to a persistent yet underexplored challenge in multilingual RAG. 

Our main contributions are as follows:
\begin{itemize}
    \item \textbf{Multilingual Dataset Construction.} We construct multilingual versions of HotpotQA, MuSiQue, and DuReader by translating and human-verifying all components (queries, answers, prompts, exemplars, and retrieved context), enabling controlled evaluation across four diverse languages.
    \item \textbf{Analysis of Language Drift.} We conduct controlled experiments that vary only the language of retrieved contexts, revealing overlooked patterns in multilingual RAG such as performance degradation, target-language inconsistency, and a strong fallback tendency to English. Chain-of-Thought traces show that drift typically arises mid-generation due to decoding-time biases.
    \item \textbf{Training-free Language Control.} We introduce SCD, a lightweight decoding-time method that softly penalizes non-target tokens. SCD is model-agnostic, requires no training, and improves both output language consistency and task accuracy across datasets and LLMs.
\end{itemize}

%We hope our findings shed light on the underexplored yet practically important phenomenon of language drift in multilingual generation. By providing both systematic evidence and a general decoding solution, we aim to inspire further research on controllable generation, cross-lingual alignment, and robust reasoning in multilingual RAG systems.

\section{Language Drift in Multilingual RAG}\label{sec.data}

In this section, we conduct a comprehensive empirical investigation into the phenomenon of language drift, where model outputs deviate from the intended target language during multilingual RAG generation. To support this study, we construct multilingual variants of several benchmark RAG datasets by translating and aligning all critical components, including queries, answers, prompts, exemplars, and retrieved passages. We then evaluate LLM behavior across a range of controlled conditions. Our findings reveal a set of systematic behaviors that undermine both task accuracy and output language alignment under cross-lingual conditions.

\subsection{Multilingual Dataset Construction}

To systematically evaluate how multilingual retrieved context influences LLM behavior in RAG, we require datasets in which the language of each input component can be independently controlled. This enables us to isolate the impact of cross-lingual retrieved passages on model reasoning and output consistency. However, no existing benchmark satisfies these constraints while remaining compatible with RAG. To address this gap, we construct multilingual versions of three widely used QA datasets that support retrieval augmentation: HotpotQA~\cite{Yang2018HotpotQAAD}, MuSiQue~\cite{trivedi2021musique}, and DuReader\footnote{\url{https://github.com/baidu/DuReader}}. These datasets contain high-quality question–answer pairs with human-annotated gold retrieved context, making them well-suited for our purpose. We select four typologically diverse languages, English (\texttt{EN}), Chinese (\texttt{ZH}), Arabic (\texttt{AR}), and Russian (\texttt{RU}), to capture a broad range of linguistic variation. Representative data examples and the format used for multilingual annotation are provided in Appendix F.

Each dataset contributes 1,000 samples. For every sample, we prepare five components: a user query, a reference answer, several gold retrieved contexts, a prompt template, and several in-context exemplars. All components are translated into the four languages using GPT-4o, followed by manual verification to ensure semantic fidelity and natural fluency. This multilingual suite enables flexible and language-controlled experimentation across a wide range of configurations. 

%This multilingual data suite allows us to precisely control the language configuration of each component and to simulate realistic multilingual RAG workflows. In later experiments, we systematically vary the language of the evidence passage while holding all other fields constant, enabling direct analysis of how cross-lingual context influences performance and output language consistency. 

\subsection{Experimental Setup}

Based on the multilingual datasets described above, we design a controlled experimental framework to evaluate how the language of the retrieved context influences output behavior in RAG. In our core setup, we fix the language of the query, prompt, and ICL examples to the target language (denoted as the context language), and vary only the language of the retrieved passage to isolate cross-lingual interference effects.

We test across three datasets (HotpotQA, MuSiQue, DuReader) and four target languages (\texttt{EN}, \texttt{ZH}, \texttt{AR}, \texttt{RU}), using two instruction-tuned LLMs as backbones: LLaMA3-8B-Instruct~\cite{grattafiori2024llama} and Qwen2.5-7B-Instruct~\cite{qwen2.5,qwen3}. All generations are performed using default decoding parameters, and each prompt includes four ICL exemplars in the same language as the query.

For evaluation, we report standard BLEU-1/2/3 and ROUGE-1/2/L scores, along with their averaged variants (\textbf{BLEU} and \textbf{ROUGE}), using reference answers in the target language as the gold standard. To further quantify language fidelity, we introduce a \textbf{Language Consistency (LC)} metric, which measures the proportion of generated responses written in the expected target language. This comprehensive metric suite allows us to jointly evaluate reasoning accuracy and language control in multilingual RAG settings.

\begin{figure}[t]
	\centering
	\includegraphics[width=0.99\linewidth]{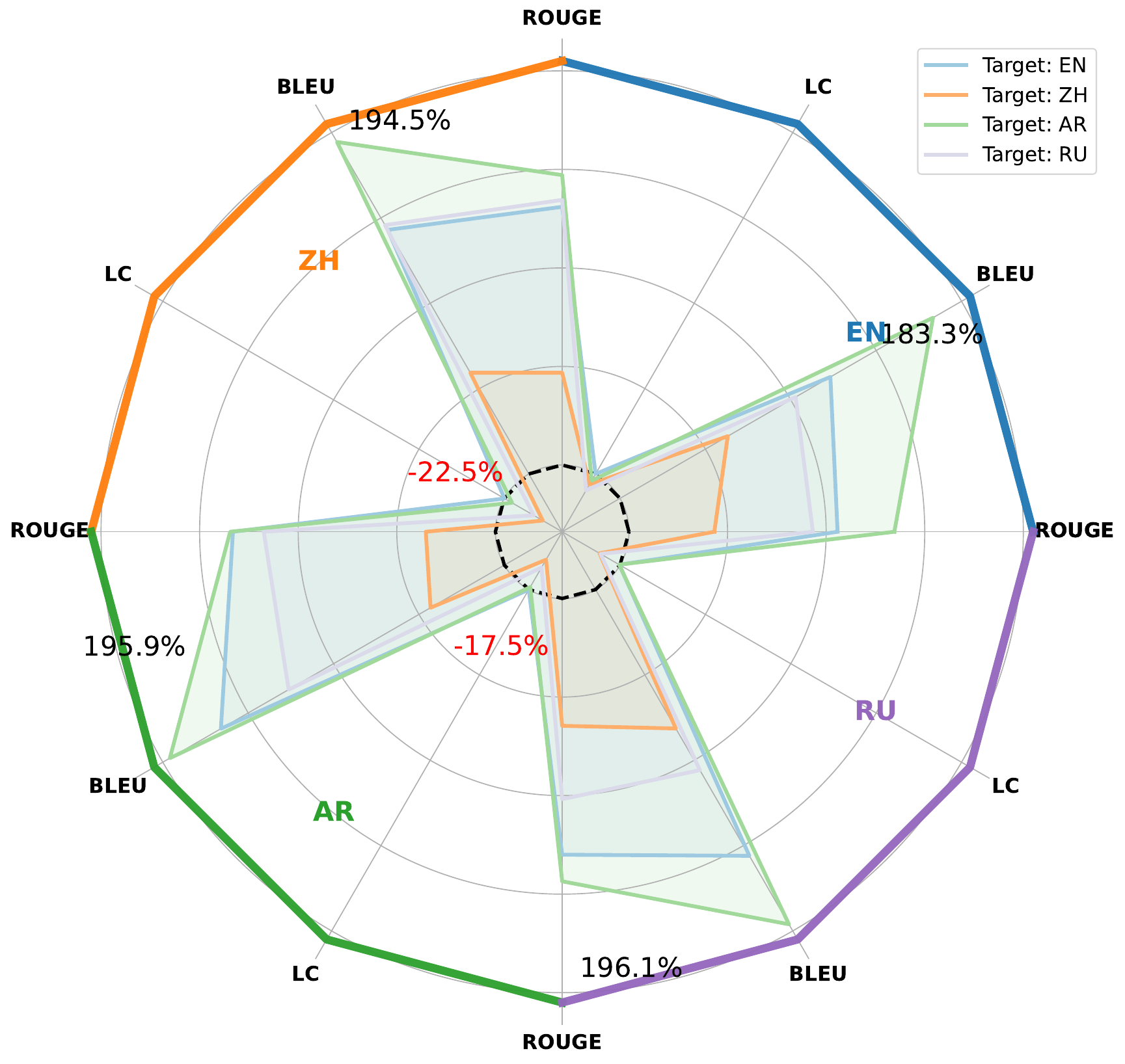}
	\caption{
Relative performance gap between with-ICL and without-ICL settings across different target and context language combinations on the HotpotQA dataset, visualized as a polar radar chart. Each axis corresponds to one evaluation metric (ROUGE, BLEU, or LC) under a specific context language (\texttt{EN}, \texttt{ZH}, \texttt{AR}, or \texttt{RU}), totaling 12 axes. Solid lines represent different target languages, color-coded accordingly. Values indicate the percentage difference between ICL and non-ICL performance under each configuration. The black dashed ring at 0\% denotes no change. Labels mark the highest gains and the most severe LC degradations. The chart reveals that ICL consistently improves BLEU and ROUGE, but often reduces language consistency, especially under \texttt{ZH} and \texttt{RU} contexts.
}

	\label{pic:radar}
\end{figure}

\subsection{ICL Improves Performance but Undermines Consistency}

To investigate how multilingual retrieved context and ICL jointly affect RAG performance, we conduct controlled experiments across various target–context language pairs. Specifically, we fix the query, prompt, and exemplars in the target language (\texttt{EN}, \texttt{ZH}, \texttt{AR}, or \texttt{RU}) and vary only the language of the retrieved context. For each configuration, we compare model outputs with and without ICL exemplars, allowing us to isolate the effects of ICL under multilingual interference.

Figure~\ref{pic:radar} summarizes these effects using a radar chart on the HotpotQA dataset with \texttt{LLaMA3-8B-Instruct} as the backbone. Each colored line represents a fixed target language, while each radial group corresponds to one context language (\texttt{EN}, \texttt{ZH}, \texttt{AR}, \texttt{RU}), covering three evaluation metrics: ROUGE, BLEU, and LC. The plotted values represent the \textit{relative percentage change} introduced by ICL compared to the non-ICL baseline under the same configuration. Positive values indicate improvements, whereas negative values reflect degradation. As a concrete example, the green point within the orange \texttt{ZH}-labeled frame represents the \texttt{ZH}-\texttt{AR} condition. It shows that ICL increases BLEU significantly but reduces LC, reflecting the common pattern where richer reasoning comes at the cost of linguistic fidelity under cross-lingual retrieved context. Due to space constraints, we report the radar plot results only for HotpotQA, which is representative of the broader trends. Similar patterns are observed across other datasets, languages, and backbone models; detailed results are included in Appendix D. Our results in Figure~\ref{pic:radar} reveal two key findings:

\begin{itemize}
    \item \textbf{Multilingual interference degrades both performance and consistency.}  
    When the retrieved context is in a language different from the target, both task performance (measured by BLEU and ROUGE) and output language consistency decline significantly. Notably, we observe that \textbf{English acts as the strongest interfering language}: when used as cross-lingual retrieved context, it induces the most severe performance degradation across non-English targets. For example, in the ICL setting with \texttt{ZH} as the target language, language consistency drops from 92.0\% to 68.4\% when switching retrieved contexts from \texttt{ZH} to \texttt{EN} retrieved context, with a drop in average BLEU score from 0.212 to 0.086. In contrast, \texttt{EN} exhibits the \textbf{strongest resistance to interference} when serving as the target language, while \texttt{ZH} shows the \textbf{greatest sensitivity} across all datasets. 

    \item \textbf{In-context learning improves performance but worsens consistency.}  
    Adding ICL examples consistently improves generation quality across all datasets and models. However, it also intensifies output language drift, leading the model to deviate further from the expected target language. For example, with \texttt{RU} as the target language, the average ROUGE increases from 0.193 to 0.373 after adding ICL, while language consistency drops from 0.991 to 0.895. Similar trends are observed when the context language differs from the target language: ICL improves accuracy but significantly reduces alignment with the expected output language.

\end{itemize}

These findings indicate that while ICL improves semantic fidelity, it also increases vulnerability to language drift due to extended reasoning and exposure to non-target-language tokens. Since ICL reflects real-world usage and consistently improves performance, we adopt it as the default in all experiments, with prompts explicitly instructing the model to generate in the target language.

\subsection{English as the Default Fallback Language}\label{sec:fallback}

While previous results show that cross-lingual interference reduces output consistency, we further investigate \textit{which language the model tends to generate} when it fails to remain in the target language. Specifically, we analyze all inconsistent outputs and identify their actual output language. Strikingly, we observe that in the majority of drift cases across all target languages and datasets, the model defaults to generating in \texttt{EN} regardless of whether the retrieved context is \texttt{EN}, as shown in Figure~\ref{pic:heatmap}. Due to the space limitations, additional results with similar conclusions are provided in the Appendix A. 

This fallback behavior suggests that \texttt{EN} plays a dominant role not only in training but also during decoding. Rather than aligning with the context language, the model often defaults to \texttt{EN} when facing ambiguity, a tendency driven by structural biases such as the over-representation of English tokens during pretraining and the concentration of factual knowledge in \texttt{EN}. Our experiments further confirm that even when both the target languages and context are non-English, misaligned outputs predominantly appear in \texttt{EN}, indicating that language drift is not random but guided by \texttt{EN} acting as a default semantic attractor.

\begin{figure}[t]
	\centering
	\includegraphics[width=0.99\linewidth]{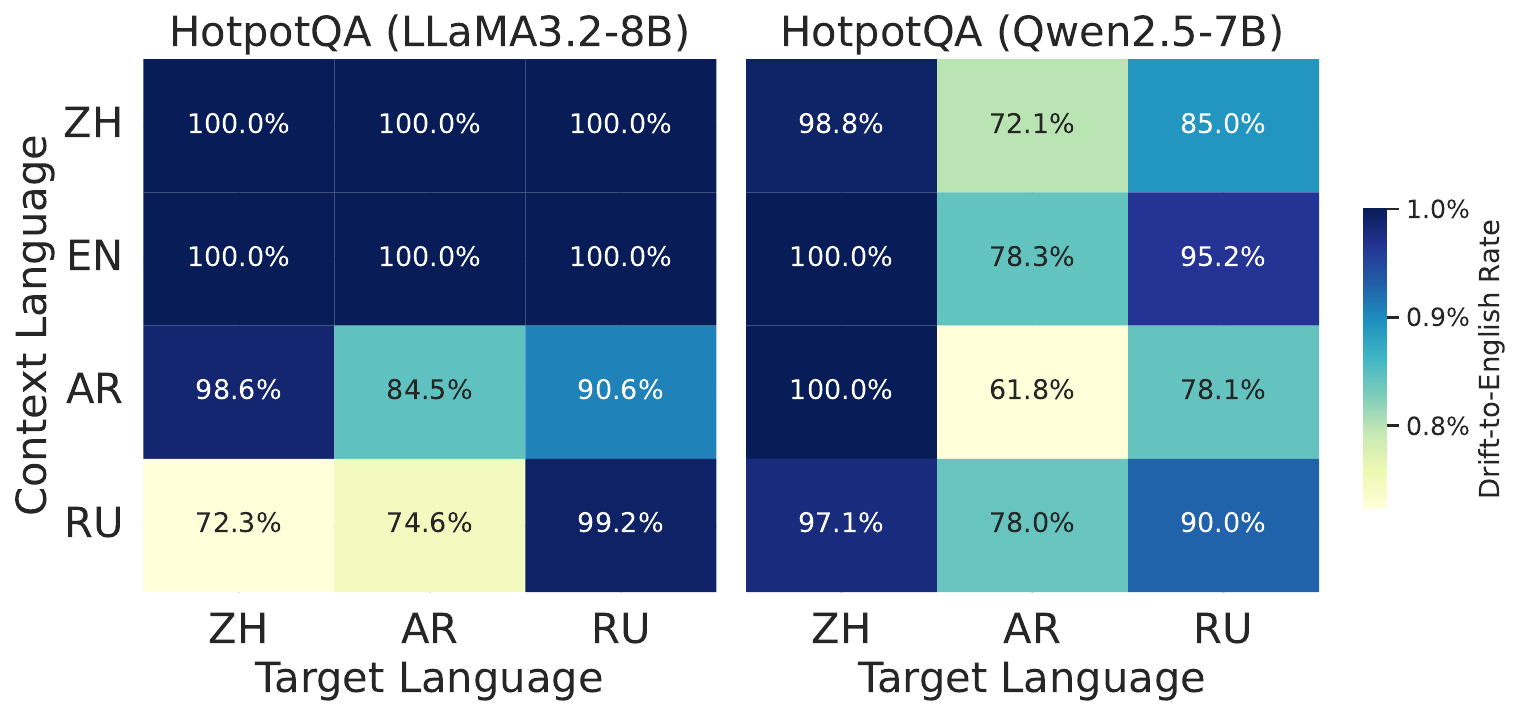}
	\caption{ Language drift patterns on the HotpotQA dataset for \texttt{LLaMA3-8B} and \texttt{Qwen2.5-7B} models. Each cell shows the percentage of inconsistent outputs that are generated in English (\texttt{EN}). Both models exhibit a strong fallback tendency toward English across all cross-lingual settings.}
	\label{pic:heatmap}
\end{figure}

\begin{table*}[]
\centering
\begin{tabular}{cc|ccc|ccc}
\bottomrule[1.5pt]
\textbf{}                                                                               & \textbf{}                                                            & \multicolumn{3}{c|}{\texttt{LLaMA3-8B}}                                                                        & \multicolumn{3}{c}{\texttt{Qwen2.5-7B}}                                                                          \\ \hline
\multicolumn{1}{c|}{\textbf{\begin{tabular}[c]{@{}c@{}}Targe \\ Language\end{tabular}}} & \textbf{\begin{tabular}[c]{@{}c@{}}Context \\ Language\end{tabular}} & \textbf{ROUGE} & \textbf{ROUGE(T)} & \textbf{\begin{tabular}[c]{@{}c@{}}Semantic \\ Match Rate\end{tabular}} & \textbf{ROUGE} & \textbf{ROUGE(T)} & \textbf{\begin{tabular}[c]{@{}c@{}}Semantic \\ Match Rate\end{tabular}} \\ \bottomrule[1.5pt]
\multicolumn{1}{c|}{\multirow{3}{*}{\texttt{ZH}}}                                       & \texttt{EN}                                                          & 0.182          & 0.263             & 54.7\%                                                                      & 0.331          & 0.352             & 62.7\%                                                                      \\
\multicolumn{1}{c|}{}                                                                   & \texttt{AR}                                                          & 0.211          & 0.258             & 46.2\%                                                                      & 0.342          & 0.366             & 55.3\%                                                                      \\
\multicolumn{1}{c|}{}                                                                   & \texttt{RU}                                                          & 0.209          & 0.261             & 49.4\%                                                                      & 0.337          & 0.359             & 54.2\%                                                                      \\ \hline
\multicolumn{1}{c|}{\multirow{3}{*}{\texttt{AR}}}                                       & \texttt{EN}                                                          & 0.294          & 0.331             & 53.4\%                                                                      & 0.201          & 0.221             & 46.4\%                                                                      \\
\multicolumn{1}{c|}{}                                                                   & \texttt{ZH}                                                          & 0.265          & 0.288             & 48.3\%                                                                      & 0.187          & 0.202             & 42.2\%                                                                      \\
\multicolumn{1}{c|}{}                                                                   & \texttt{RU}                                                          & 0.280          & 0.303             & 50.0\%                                                                      & 0.206          & 0.220             & 43.0\%                                                                      \\ \hline
\multicolumn{1}{c|}{\multirow{3}{*}{\texttt{RU}}}                                       & \texttt{EN}                                                          & 0.333          & 0.388             & 62.8\%                                                                      & 0.240          & 0.262             & 60.1\%                                                                      \\
\multicolumn{1}{c|}{}                                                                   & \texttt{ZH}                                                          & 0.335          & 0.367             & 59.1\%                                                                      & 0.248          & 0.257             & 56.1\%                                                                      \\
\multicolumn{1}{c|}{}                                                                   & \texttt{AR}                                                          & 0.339          & 0.361             & 62.9\%                                                                      & 0.248          & 0.252             & 59.0\%                                                                      \\ \bottomrule[1.5pt]
\end{tabular}
\caption{Performance under cross-lingual retrieved context for non-English target languages (\texttt{ZH}, \texttt{AR}, \texttt{RU}) using \texttt{LLaMA3-8B} and \texttt{Qwen2.5-7B} on HotpotQA. We report standard ROUGE, ROUGE after translating the model output to the target language (ROUGE(T)), and Semantic Match Rate assessed by GPT. Despite language drift, many outputs remain semantically correct, highlighting decoder-level instability rather than comprehension failure.}
\end{table*}\label{tab:collapse}

\subsection{Language Collapse During Decoding}

To assess whether the observed language drift arises from comprehension failure or unstable decoding behavior, we conduct a semantic agreement analysis. As shown in Table~1, we compare three evaluation metrics under cross-lingual settings:  (1) Standard ROUGE between the model output and the target-language reference;  (2) ROUGE after translating drifted outputs back into the target language and recomputing scores against the original reference (denoted as ROUGE(T));  (3) Semantic Match Rate, scored by GPT-4o, which evaluates whether the model output is factually aligned with the reference regardless of surface language. We observe that translation leads to a \textit{significant} improvement in ROUGE scores. For example, ROUGE increases from 0.182 to 0.263 for \texttt{ZH}, and from 0.333 to 0.388 for \texttt{RU}, indicating that the original outputs are semantically aligned despite being expressed in the wrong language. Moreover, the Semantic Match Rate further confirms that even when ROUGE is low, the match rate often exceeds 60\% for \texttt{RU} and over 50\% for \texttt{ZH} and \texttt{AR}, demonstrating strong task understanding. These findings suggest that language drift stems not from comprehension failure but from decoder-level instability. Additional results on other datasets (see Appendix~B) show similar patterns across models and languages.

This pattern suggests a language collapse during decoding, where the LLM correctly processes the input and understands the intended task but fails to maintain the target language throughout generation. We hypothesize that this issue arises from token-level priors learned during pretraining, as English tokens tend to dominate due to their higher frequency, more stable syntactic structures, and richer factual coverage. During multi-step reasoning, especially under CoT prompting, such biases can override explicit language instructions and gradually shift the generation toward English. The drift typically unfolds over time, with the generation beginning in the target language but progressively deviating into English. This highlights a fundamental limitation in multilingual LLMs: \textbf{strong semantic reasoning does not guarantee stable language control during generation.}

\begin{figure}[H]
	\centering
	\includegraphics[width=0.99\linewidth]{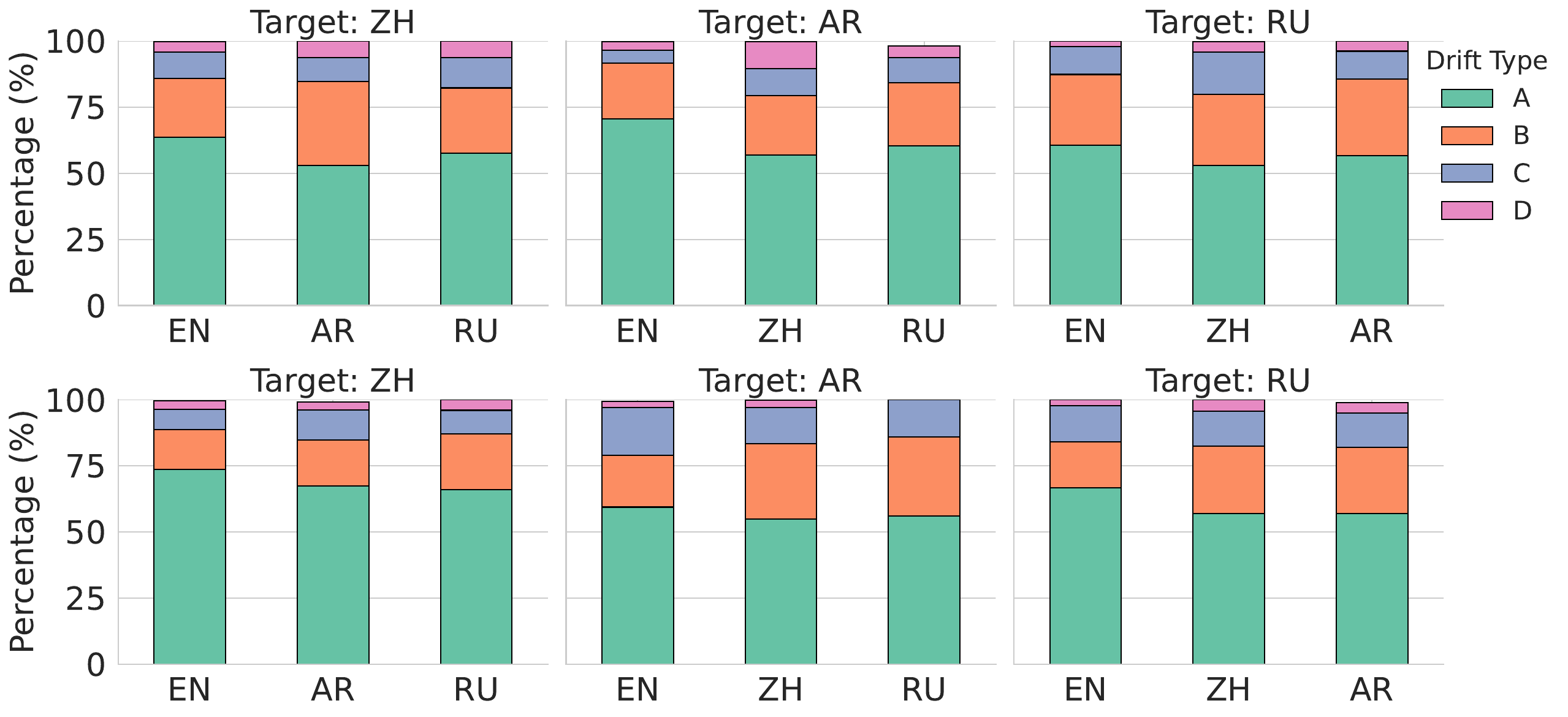}
	\caption{Distribution of four language drift types across different target–context language pairs in the HotpotQA dataset. Each subplot corresponds to a fixed target language (\texttt{ZH}, \texttt{AR}, \texttt{RU}), with the x-axis denoting the context language. The top row displays results for \texttt{LLaMA3-8B}, and the bottom row for \texttt{Qwen2.5-7B}. }
	\label{pic:drift_type}
\end{figure}

\subsection{Types of Language Drift Behaviors}
To better understand how language drift manifests in multilingual reasoning, we categorize drifted outputs into four distinct behavioral types based on multilingual generations. We randomly sampled 1,000 language-inconsistent outputs and had them manually annotated by three trained reviewers with backgrounds in linguistics or multilingual NLP. The taxonomy includes: \textbf{Type A:}  \textit{Named Entity Representation Divergence}, where inconsistent transliteration or spelling results in mismatches despite semantic equivalence; \textbf{Type B:} \textit{Answer Target Shift}, where the model alters answer granularity or is misled by context-language cues, leading to an incorrect sub-answer; \textbf{Type C:} \textit{Reasoning Chain Misalignment}, where the CoT path becomes structurally disrupted due to language mixing or code-switching; and \textbf{Type D:} \textit{Conceptual Reference Shift}, where cultural or semantic biases embedded in the dominant language (such as English) trigger unintended knowledge concepts. Full category definitions and examples are provided in Appendix C.

We use GPT-4o to classify a representative set of drifted outputs into the four categories defined in our taxonomy, followed by manual verification to ensure label quality. As shown in Figure~\ref{pic:drift_type}, the most common behavior across both models and all target languages is \textit{Named Entity Representation Divergence} (Type A), which accounts for approximately 55\% to 74\% of drifted cases on average. This is followed by \textit{Answer Target Shift} (Type B), occurring in roughly 17\% to 31\% of cases, with greater variation across context languages. \textit{Reasoning Chain Misalignment} (Type C) is less frequent, comprising around 9\% to 18\%, while \textit{Conceptual Reference Shift} (Type D) remains rare, often below 5\%.

These findings suggest that most drift cases arise from surface-level inconsistencies, such as entity formatting or answer phrasing, rather than from deeper reasoning failures. Recognizing how such drift emerges during the later stages of CoT decoding can inform more targeted control strategies, including applying penalties for answer-level deviations or reinforcing consistency in entity representation.

\begin{table*}[t]
\begin{tabular}{ccccccccccc}
\bottomrule[1.5pt]
\textbf{}                                                                               & \multicolumn{1}{c|}{\textbf{}}                                                            & \multicolumn{3}{c|}{\textbf{HotpotQA}}                            & \multicolumn{3}{c|}{\textbf{Musique}}                             & \multicolumn{3}{c}{\textbf{DuReader}}        \\ \hline
\multicolumn{1}{c|}{\textbf{\begin{tabular}[c]{@{}c@{}}Targe \\ Language\end{tabular}}} & \multicolumn{1}{c|}{\textbf{\begin{tabular}[c]{@{}c@{}}Context \\ Language\end{tabular}}} & \textbf{ROUGE} & \textbf{BLEU} & \multicolumn{1}{c|}{\textbf{LC}} & \textbf{ROUGE} & \textbf{BLEU} & \multicolumn{1}{c|}{\textbf{LC}} & \textbf{ROUGE} & \textbf{BLEU} & \textbf{LC} \\ \bottomrule[1.5pt]
\multicolumn{11}{l}{\textit{\textbf{Prompted Language Instruction}}}                                                                                                                                                                                                                                                                                                       \\ \hline
\multicolumn{1}{c|}{\multirow{3}{*}{\texttt{ZH}}}                                       & \multicolumn{1}{c|}{\texttt{EN}}                                                          & 0.182          & 0.086         & \multicolumn{1}{c|}{68.4\%}      & 0.187          & 0.097         & \multicolumn{1}{c|}{63.9\%}      & 0.339          & 0.166         & 84.2\%      \\
\multicolumn{1}{c|}{}                                                                   & \multicolumn{1}{c|}{\texttt{AR}}                                                          & 0.211          & 0.106         & \multicolumn{1}{c|}{77.7\%}      & 0.181          & 0.089         & \multicolumn{1}{c|}{76.5\%}      & 0.358          & 0.175         & 90.1\%      \\
\multicolumn{1}{c|}{}                                                                   & \multicolumn{1}{c|}{\texttt{RU}}                                                          & 0.209          & 0.107         & \multicolumn{1}{c|}{79.5\%}      & 0.169          & 0.087         & \multicolumn{1}{c|}{64.5\%}      & 0.343          & 0.168         & 83.1\%      \\ \hline
\multicolumn{1}{c|}{\multirow{3}{*}{\texttt{AR}}}                                       & \multicolumn{1}{c|}{\texttt{EN}}                                                          & 0.294          & 0.162         & \multicolumn{1}{c|}{85.4\%}      & 0.144          & 0.080         & \multicolumn{1}{c|}{90.0\%}      & 0.209          & 0.099         & 88.2\%      \\
\multicolumn{1}{c|}{}                                                                   & \multicolumn{1}{c|}{\texttt{ZH}}                                                          & 0.265          & 0.143         & \multicolumn{1}{c|}{88.4\%}      & 0.120          & 0.057         & \multicolumn{1}{c|}{89.2\%}      & 0.193          & 0.080         & 87.0\%      \\
\multicolumn{1}{c|}{}                                                                   & \multicolumn{1}{c|}{\texttt{RU}}                                                          & 0.280          & 0.151         & \multicolumn{1}{c|}{88.6\%}      & 0.121          & 0.061         & \multicolumn{1}{c|}{89.8\%}      & 0.186          & 0.077         & 89.5\%      \\ \hline
\multicolumn{1}{c|}{\multirow{3}{*}{\texttt{RU}}}                                       & \multicolumn{1}{c|}{\texttt{EN}}                                                          & 0.333          & 0.177         & \multicolumn{1}{c|}{80.2\%}      & 0.218          & 0.119         & \multicolumn{1}{c|}{81.9\%}      & 0.285          & 0.150         & 84.3\%      \\
\multicolumn{1}{c|}{}                                                                   & \multicolumn{1}{c|}{\texttt{ZH}}                                                          & 0.335          & 0.172         & \multicolumn{1}{c|}{85.1\%}      & 0.206          & 0.102         & \multicolumn{1}{c|}{90.2\%}      & 0.296          & 0.149         & 85.8\%      \\
\multicolumn{1}{c|}{}                                                                   & \multicolumn{1}{c|}{\texttt{AR}}                                                          & 0.339          & 0.179         & \multicolumn{1}{c|}{86.8\%}      & 0.214          & 0.109         & \multicolumn{1}{c|}{92.5\%}      & 0.288          & 0.143         & 90.9\%      \\ \hline
\multicolumn{11}{l}{\textit{\textbf{Translation-Based Evaluation}}}                                                                                                                                                                                                                                                                                                        \\ \hline
\multicolumn{1}{c|}{\multirow{3}{*}{\texttt{ZH}}}                                       & \multicolumn{1}{c|}{\texttt{EN}}                                                          & 0.263          & 0.135         & \multicolumn{1}{c|}{100.0\%}     & 0.257          & 0.142         & \multicolumn{1}{c|}{100.0\%}     & 0.366          & 0.178         & 100.0\%     \\
\multicolumn{1}{c|}{}                                                                   & \multicolumn{1}{c|}{\texttt{AR}}                                                          & 0.258          & 0.132         & \multicolumn{1}{c|}{100.0\%}     & 0.214          & 0.105         & \multicolumn{1}{c|}{100.0\%}     & 0.364          & 0.177         & 100.0\%     \\
\multicolumn{1}{c|}{}                                                                   & \multicolumn{1}{c|}{\texttt{RU}}                                                          & 0.261          & 0.136         & \multicolumn{1}{c|}{100.0\%}     & 0.235          & 0.124         & \multicolumn{1}{c|}{100.0\%}     & 0.365          & 0.175         & 100.0\%     \\ \hline
\multicolumn{1}{c|}{\multirow{3}{*}{\texttt{AR}}}                                       & \multicolumn{1}{c|}{\texttt{EN}}                                                          & 0.331          & 0.183         & \multicolumn{1}{c|}{100.0\%}     & 0.168          & 0.095         & \multicolumn{1}{c|}{100.0\%}     & 0.231          & 0.114         & 100.0\%     \\
\multicolumn{1}{c|}{}                                                                   & \multicolumn{1}{c|}{\texttt{ZH}}                                                          & 0.288          & 0.156         & \multicolumn{1}{c|}{100.0\%}     & 0.135          & 0.066         & \multicolumn{1}{c|}{100.0\%}     & 0.202          & 0.087         & 100.0\%     \\
\multicolumn{1}{c|}{}                                                                   & \multicolumn{1}{c|}{\texttt{RU}}                                                          & 0.303          & 0.165         & \multicolumn{1}{c|}{100.0\%}     & 0.140          & 0.074         & \multicolumn{1}{c|}{100.0\%}     & 0.195          & 0.083         & 100.0\%     \\ \hline
\multicolumn{1}{c|}{\multirow{3}{*}{\texttt{RU}}}                                       & \multicolumn{1}{c|}{\texttt{EN}}                                                          & 0.388          & 0.218         & \multicolumn{1}{c|}{100.0\%}     & 0.258          & 0.148         & \multicolumn{1}{c|}{100.0\%}     & 0.314          & 0.167         & 100.0\%     \\
\multicolumn{1}{c|}{}                                                                   & \multicolumn{1}{c|}{\texttt{ZH}}                                                          & 0.367          & 0.196         & \multicolumn{1}{c|}{100.0\%}     & 0.215          & 0.109         & \multicolumn{1}{c|}{100.0\%}     & 0.309          & 0.156         & 100.0\%     \\
\multicolumn{1}{c|}{}                                                                   & \multicolumn{1}{c|}{\texttt{AR}}                                                          & 0.361          & 0.196         & \multicolumn{1}{c|}{100.0\%}     & 0.217          & 0.114         & \multicolumn{1}{c|}{100.0\%}     & 0.293          & 0.148         & 100.0\%     \\ \hline
\multicolumn{11}{l}{\textit{\textbf{Soft-Constrained Decoding (Ours)}}}                                                                                                                                                                                                                                                                                                    \\ \hline
\multicolumn{1}{c|}{\multirow{3}{*}{\texttt{ZH}}}                                       & \multicolumn{1}{c|}{\texttt{EN}}                                                          & 0.306          & 0.155         & \multicolumn{1}{c|}{90.6\%}      & 0.276          & 0.146         & \multicolumn{1}{c|}{91.8\%}      & 0.403          & 0.190         & 95.2\%      \\
\multicolumn{1}{c|}{}                                                                   & \multicolumn{1}{c|}{\texttt{AR}}                                                          & 0.283          & 0.146         & \multicolumn{1}{c|}{93.9\%}      & 0.234          & 0.118         & \multicolumn{1}{c|}{94.8\%}      & 0.408          & 0.195         & 96.6\%      \\
\multicolumn{1}{c|}{}                                                                   & \multicolumn{1}{c|}{\texttt{RU}}                                                          & 0.293          & 0.156         & \multicolumn{1}{c|}{92.5\%}      & 0.243          & 0.130         & \multicolumn{1}{c|}{92.3\%}      & 0.404          & 0.190         & 95.7\%      \\ \hline
\multicolumn{1}{c|}{\multirow{3}{*}{\texttt{AR}}}                                       & \multicolumn{1}{c|}{\texttt{EN}}                                                          & 0.352          & 0.197         & \multicolumn{1}{c|}{96.4\%}      & 0.187          & 0.106         & \multicolumn{1}{c|}{98.8\%}      & 0.241          & 0.113         & 96.7\%      \\
\multicolumn{1}{c|}{}                                                                   & \multicolumn{1}{c|}{\texttt{ZH}}                                                          & 0.312          & 0.170         & \multicolumn{1}{c|}{95.5\%}      & 0.157          & 0.079         & \multicolumn{1}{c|}{97.6\%}      & 0.236          & 0.104         & 94.1\%      \\
\multicolumn{1}{c|}{}                                                                   & \multicolumn{1}{c|}{\texttt{RU}}                                                          & 0.326          & 0.183         & \multicolumn{1}{c|}{96.3\%}      & 0.152          & 0.080         & \multicolumn{1}{c|}{98.0\%}      & 0.220          & 0.092         & 95.4\%      \\ \hline
\multicolumn{1}{c|}{\multirow{3}{*}{\texttt{RU}}}                                       & \multicolumn{1}{c|}{\texttt{EN}}                                                          & 0.422          & 0.238         & \multicolumn{1}{c|}{95.4\%}      & 0.270          & 0.162         & \multicolumn{1}{c|}{94.1\%}      & 0.334          & 0.174         & 94.4\%      \\
\multicolumn{1}{c|}{}                                                                   & \multicolumn{1}{c|}{\texttt{ZH}}                                                          & 0.400          & 0.216         & \multicolumn{1}{c|}{94.1\%}      & 0.230          & 0.126         & \multicolumn{1}{c|}{94.7\%}      & 0.335          & 0.165         & 94.3\%      \\
\multicolumn{1}{c|}{}                                                                   & \multicolumn{1}{c|}{\texttt{AR}}                                                          & 0.392          & 0.216         & \multicolumn{1}{c|}{94.0\%}      & 0.232          & 0.128         & \multicolumn{1}{c|}{94.3\%}      & 0.317          & 0.155         & 94.7\%      \\ \bottomrule[1.5pt]
\end{tabular}
\caption{Performance comparison across three language control strategies: Prompted Language Instruction, Translation-Based Evaluation, and SCD on three multilingual RAG datasets. We report results for \texttt{LLaMA3-8B}, where SCD consistently improves both LC and content metrics across datasets compared to strong baselines. Results for \texttt{Qwen2.5-7B} are provided in Appendix E due to space constraints.}
\end{table*}\label{tab:main_results}

\section{Soft-Constrained Decoding}

\subsection{Soft-Constrained Decoding (SCD)}

To mitigate output language drift in multilingual generation, we propose \textbf{Soft-Constrained Decoding (SCD)}, a lightweight decoding-time control strategy that incorporates token-level language awareness into the generation process. Instead of applying rigid vocabulary restrictions, SCD subtly adjusts the token probability distribution to favor the target language, while preserving open-ended reasoning capabilities and fluent output.

\paragraph{Token Categorization.}
Let \( \mathcal{V} \) denote the model vocabulary, and we partition \( \mathcal{V} \) into three disjoint sets:
\begin{itemize}
    \item \( \mathcal{V}_{\text{target}} \): tokens associated with the \textit{target language},
    \item \( \mathcal{V}_{\text{neutral}} \): \textit{neutral tokens} such as punctuation, digits, and shared symbols,
    \item \( \mathcal{V}_{\text{distractor}} \): tokens linked to \textit{non-target languages}.
\end{itemize}
This categorization is performed via Unicode ranges or tokenizer-based heuristics and cached prior to generation.

\paragraph{Logits Adjustment.}
Let \( \mathbf{z}^{(t)} \in \mathbb{R}^{|\mathcal{V}|} \) be the raw logits output at decoding step \( t \). SCD adjusts \( \mathbf{z}^{(t)} \) before softmax as follows:
\[
\tilde{z}^{(t)}_i =
\begin{cases}
\alpha z^{(t)}_i, & \text{if } i \in \mathcal{V}_{\text{target}} \\
z^{(t)}_i, & \text{if } i \in \mathcal{V}_{\text{neutral}} \\
\beta z^{(t)}_i, & \text{if } i \in \mathcal{V}_{\text{distractor}}
\end{cases}
\]
Here, \( \alpha > 1.0 \) is a soft boost to target-language tokens, and \( \beta < 1.0 \) is a penalty for distractor-language tokens. This modification biases generation while preserving flexibility.

\paragraph{Cold Start Smoothing.}
Multilingual LLMs, especially in low-resource languages, often generate unstable initial outputs such as repeated prompts or template fragments. To minimize such disruptions, we introduce a \textit{warm-up period} by delaying the activation of language constraints until decoding step \( T_{\text{start}} \). This design ensures a fluent transition into reasoning before language control is applied.

\paragraph{Integration.}
SCD is \textit{model-agnostic} and fully compatible with standard decoding algorithms. It requires no additional training or architectural changes.

SCD operates as a lightweight decoding-time strategy that gently discourages the selection of non-target language tokens without eliminating them entirely. By incorporating language awareness directly into the token selection process, SCD guides the model to favor tokens in the target language while retaining the flexibility needed for open-ended reasoning.
 %The key idea is to dynamically down-weight distractor-language tokens in the final vocabulary distribution, thus shaping the model's trajectory without overriding its generative capacity.

\subsection{Experimental Setup and Baselines}
We evaluate our proposed SCD on three multilingual retrieval-augmented QA datasets, i.e., HotpotQA, MuSiQue, and DuReader, which are described in Section 2.1. Experiments are conducted using two instruction-tuned LLMs: LLaMA3-8B-Instruct and Qwen2.5-7B-Instruct. We empirically find moderate settings (\( \alpha = 1.1 \), \( \beta = 0.9 \), \( T_{\text{start}} = 5 \)) to balance language fidelity and semantic fluency in SCD.

To benchmark SCD against other lightweight language control strategies, we compare it with the following decoding-time baselines: (1) \textbf{Prompted Language Instruction}: Explicitly appending an instruction in the prompt that requests answers to be generated in the target language; (2) \textbf{Translation-Based Evaluation}: Evaluating drifted outputs by translating them back into the target language using the same LLM, before computing BLEU/ROUGE scores; (3) \textbf{Vocabulary Restriction Decoding}: Restricting the decoding space to tokens belonging to the target language only, effectively applying a hard constraint on generation.

We evaluate all methods using three complementary metrics: (1) BLEU (mean of BLEU-1/2/3), (2) ROUGE (mean of ROUGE-1/2/L), and (3) language consistency (LC), defined as the percentage of outputs generated in the correct target language. All decoding parameters follow the default settings of each model, and no task-specific or model-specific fine-tuning is applied. Additional performance improvements may be obtained by tuning decoding parameters, we leave this for future work. All reported scores are averaged over five independent runs to reduce randomness.

\subsection{Effectiveness of Soft-Constrained Decoding}

As shown in Table 2, SCD consistently outperforms existing language control methods, achieving notable improvements in both \textit{language consistency} (LC) and \textit{semantic generation quality}, as measured by average BLEU and ROUGE scores. These results support our central hypothesis that maintaining alignment with the target language can reinforce, rather than hinder, the coherence and accuracy of reasoning paths.

Across all datasets and language configurations, SCD consistently improves both language consistency and content quality compared to the Prompted Language Instruction baseline. For instance, under the challenging \texttt{ZH-EN} condition on HotpotQA, SCD increases LC from 68.4\% to 90.6\%, while also boosting BLEU from 0.086 to 0.155 and ROUGE from 0.182 to 0.306. Similar trends are observed for other target languages such as \texttt{AR} and \texttt{RU}, with LC improvements ranging from 10 to 22 percentage points.

While the translation-based method trivially achieves 100\% LC by converting drifted outputs into the target language after generation, it often underperforms SCD in BLEU and ROUGE. This outcome is expected, as translation does not recover the original reasoning trajectory but merely reformulates its surface form. Moreover, translation-based evaluation adds additional complexity, increases inference cost, and may amplify noise when the original outputs are incomplete or syntactically broken. 

The above results demonstrate that SCD is a practical, lightweight, and model-agnostic decoding-time intervention. It requires no additional training or architectural modifications, and can be seamlessly integrated into standard decoding workflows (e.g., greedy, sampling, top-$p$). Across models, languages, and datasets, SCD provides consistent and substantial improvements in both linguistic alignment and semantic quality, making it a strong candidate for real-world multilingual RAG and generation-based applications.

% Please add the following required packages to your document preamble:
% \usepackage{multirow}
\begin{table}[]
\setlength{\tabcolsep}{1.0mm}
\begin{tabular}{cc|ccc|ccc}
\hline
\multicolumn{2}{c|}{\textbf{}}                                                                                                                           & \multicolumn{3}{c|}{\textbf{ROUGE}}       & \multicolumn{3}{c}{\textbf{CoT Length}}   \\ \hline
\multicolumn{1}{c|}{\textbf{\begin{tabular}[c]{@{}c@{}}Targe \\ Lang.\end{tabular}}} & \textbf{\begin{tabular}[c]{@{}c@{}}Context \\ Lang.\end{tabular}} & \textbf{PLI} & \textbf{VRD} & \textbf{SCD} & \textbf{PLI} & \textbf{VRD} & \textbf{SCD} \\ \hline
\multicolumn{1}{c|}{\multirow{3}{*}{\texttt{ZH}}}                                    & \texttt{EN}                                                       & 0.182       & 0.155        & 0.306        & 104.0       & 38.6         & 134.9        \\
\multicolumn{1}{c|}{}                                                                & \texttt{AR}                                                       & 0.211       & 0.184        & 0.283        & 103.5       & 40.2         & 142.4        \\
\multicolumn{1}{c|}{}                                                                & \texttt{RU}                                                       & 0.209       & 0.173        & 0.293        & 77.6        & 42.8         & 143.1        \\ \hline
\multicolumn{1}{c|}{\multirow{3}{*}{\texttt{AR}}}                                    & \texttt{EN}                                                       & 0.294       & 0.295        & 0.352        & 77.0        & 50.5         & 90.2         \\
\multicolumn{1}{c|}{}                                                                & \texttt{ZH}                                                       & 0.265       & 0.266        & 0.312        & 86.4        & 49.9         & 92.9         \\
\multicolumn{1}{c|}{}                                                                & \texttt{RU}                                                       & 0.280       & 0.281        & 0.326        & 86.3        & 57.4         & 100.2        \\ \hline
\multicolumn{1}{c|}{\multirow{3}{*}{\texttt{RU}}}                                    & \texttt{EN}                                                       & 0.333       & 0.343        & 0.422        & 85.6        & 58.8         & 111.8        \\
\multicolumn{1}{c|}{}                                                                & \texttt{ZH}                                                       & 0.335       & 0.339        & 0.400        & 89.4        & 56.1         & 111.0        \\
\multicolumn{1}{c|}{}                                                                & \texttt{AR}                                                       & 0.339       & 0.341        & 0.392        & 89.4        & 56.6         & 111.8        \\ \hline
\end{tabular}
\caption{
Comparison of three decoding strategies on HotpotQA across ROUGE score and average CoT length.}
\end{table}\label{tab:po-vrd-scd}
%SCD consistently achieves the best balance between semantic accuracy and reasoning depth, outperforming both baselines in ROUGE while maintaining long-form inference.

\subsection{Should Multilingual Generation Be Fully Language Isolated?}

To examine the trade-offs of different language control strategies in multilingual generation, we compare three decoding methods: Prompted Language Instruction (PLI), Vocabulary-Restricted Decoding (VRD), and our proposed SCD. PLI uses explicit prompts to enforce the target language; VRD imposes hard constraints by restricting generation to target-language tokens; and SCD softly penalizes non-target tokens while maintaining generation flexibility.

As shown in Table 3, SCD consistently achieves the highest ROUGE scores across all target–context language pairs on HotpotQA. For instance, in the \texttt{ZH}-\texttt{EN} setting, SCD reaches 0.306 ROUGE, compared to 0.155 under VRD and 0.182 under PLI. Similar trends are observed for \texttt{AR} and \texttt{RU} targets. Interestingly, VRD often underperforms PLI, suggesting that overly strict language filtering can suppress useful multilingual cues and degrade output quality, despite improving consistency.

To assess generation dynamics, we compare the average length of generated CoT responses. VRD consistently yields the shortest outputs, e.g., only 38.6 tokens in \texttt{ZH}-\texttt{EN}, compared to 104.0 with PLI and 134.9 with SCD—indicating that hard constraints truncate reasoning. In contrast, SCD preserves longer and more complete reasoning chains by allowing controlled cross-lingual flexibility. We further analyze how reasoning length affects language drift and control effectiveness in Appendix G, where SCD demonstrates robust performance across various CoT trajectories.

These results suggest that \textit{effective multilingual generation does not require full language isolation}. Allowing limited access to non-target tokens during reasoning, while softly guiding the output toward the desired language, improves both language consistency and semantic fidelity.

\section{Related Work}

Multilingual RAG has received increasing attention as a means to enhance LLMs with access to cross-lingual knowledge. Prior research has primarily focused on improving the quality of multilingual retrieval~\cite{Liu2025XRAGCR, Chirkova2024RetrievalaugmentedGI, Ranaldi2025MultilingualRG}, aligning retrieved passages with user queries across languages~\cite{Ranaldi2025ImprovingMR, Blandon2025MEMERAGAM}, and adapting RAG pipelines to typologically diverse settings~\cite{Wu2024NotAL, Zeng2024EnglishLH}. These efforts have significantly advanced retrieval-stage effectiveness in non-English tasks and established multilingual evaluation protocols. Some recent works have further explored language preferences in RAG models~\cite{Park2025InvestigatingLP,Shi2022LanguageMA, Yu2025CrossLingualCA}, highlighting accuracy disparities across languages. However, most of these works either evaluate generation outcomes at the answer level or focus on upstream retrieval modules, without deeply investigating how language behavior evolves throughout the decoding process.

In contrast, we focus on the overlooked issue of \textit{language drift} in multilingual RAG, where model outputs shift away from the target language during reasoning. We demonstrate that this drift arises during decoding, with English acting as a default fallback. To mitigate it, we propose a lightweight decoding-time strategy that improves language alignment without requiring model retraining.

\section{Conclusion}
This work addresses a key challenge in multilingual RAG: large language models often generate outputs in unintended languages when reasoning over cross-lingual evidence. Through controlled experiments and CoT analysis, we find that such drift arises from decoder-stage biases rather than comprehension failure. To mitigate this, we introduce SCD, a lightweight, model-agnostic strategy that softly penalizes non-target-language tokens. SCD consistently enhances both language consistency and task performance across models, languages, and datasets. These findings underscore the value of decoding-time control for building more robust and controllable multilingual RAG systems.

\section*{Acknowledgements}
This work was supported by the National Natural Science Foundation of China (Grant No. 62276089), the Natural Science Foundation of Tianjin (Grant No. 24JCJQJC00200 and Grant No. 24JCQNJC01230), the Natural Science Foundation of Hebei Province (Grant No. F2024202064), the Science Research Project of Hebei Education Department (Grant No. BJ2025004), the Ministry of Human Resources and Social Security of China (Grant No. RSTH-2023-135-1), and the Science and Technology Program of Hebei Province (Grant No. 24464401D).

\bibliography{aaai2026}

\clearpage

\subsection*{Appendix A: Supplementary Fallback Analysis on Musique and DuReader}\label{app:fallback}

\begin{figure}[h]
	\centering
	\includegraphics[width=0.99\linewidth]{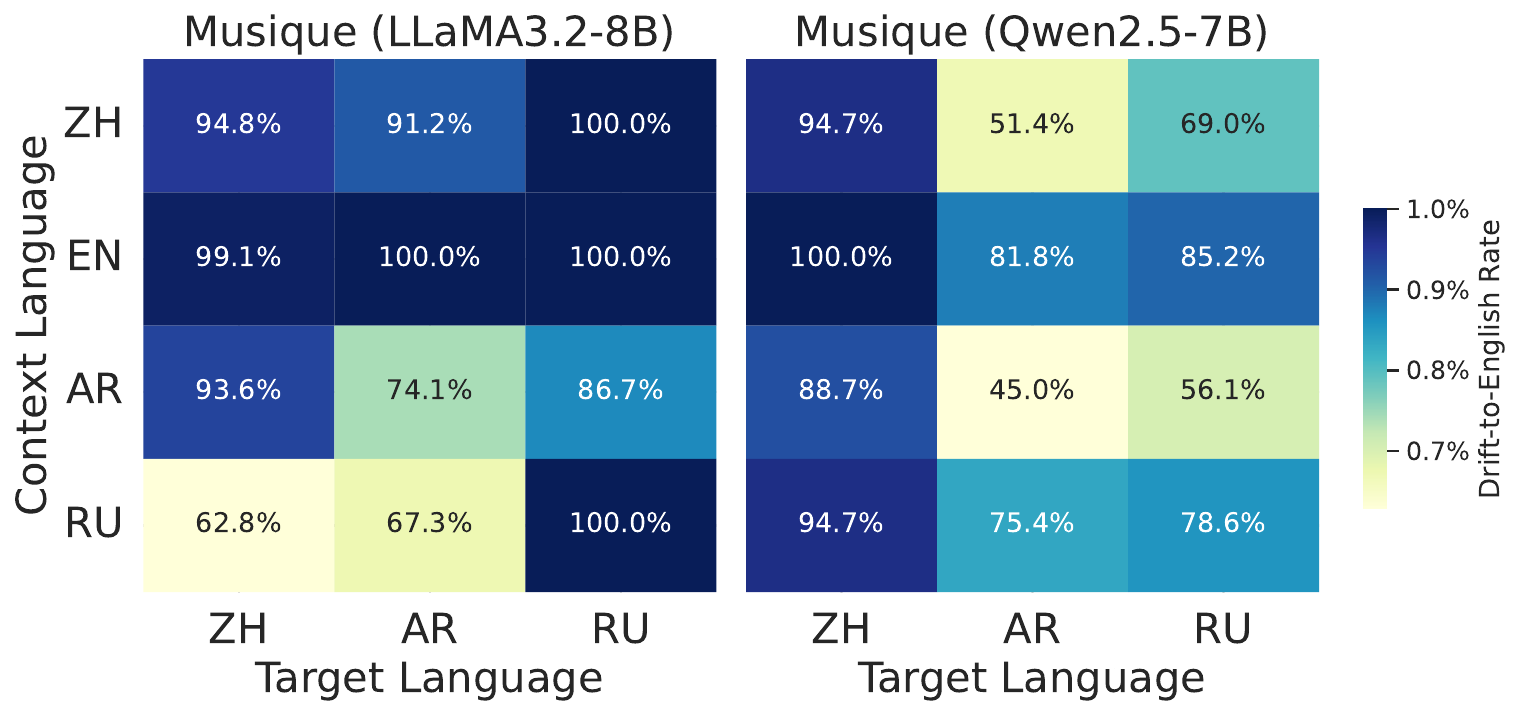}
	\caption{ Language drift patterns on the Musique dataset for LLaMA3-8B and Qwen2.5-7B models. Each cell shows the percentage of inconsistent outputs that are generated in English. }
	\label{pic:heatmap_musique}
\end{figure}

\begin{figure}[h]
	\centering
	\includegraphics[width=0.99\linewidth]{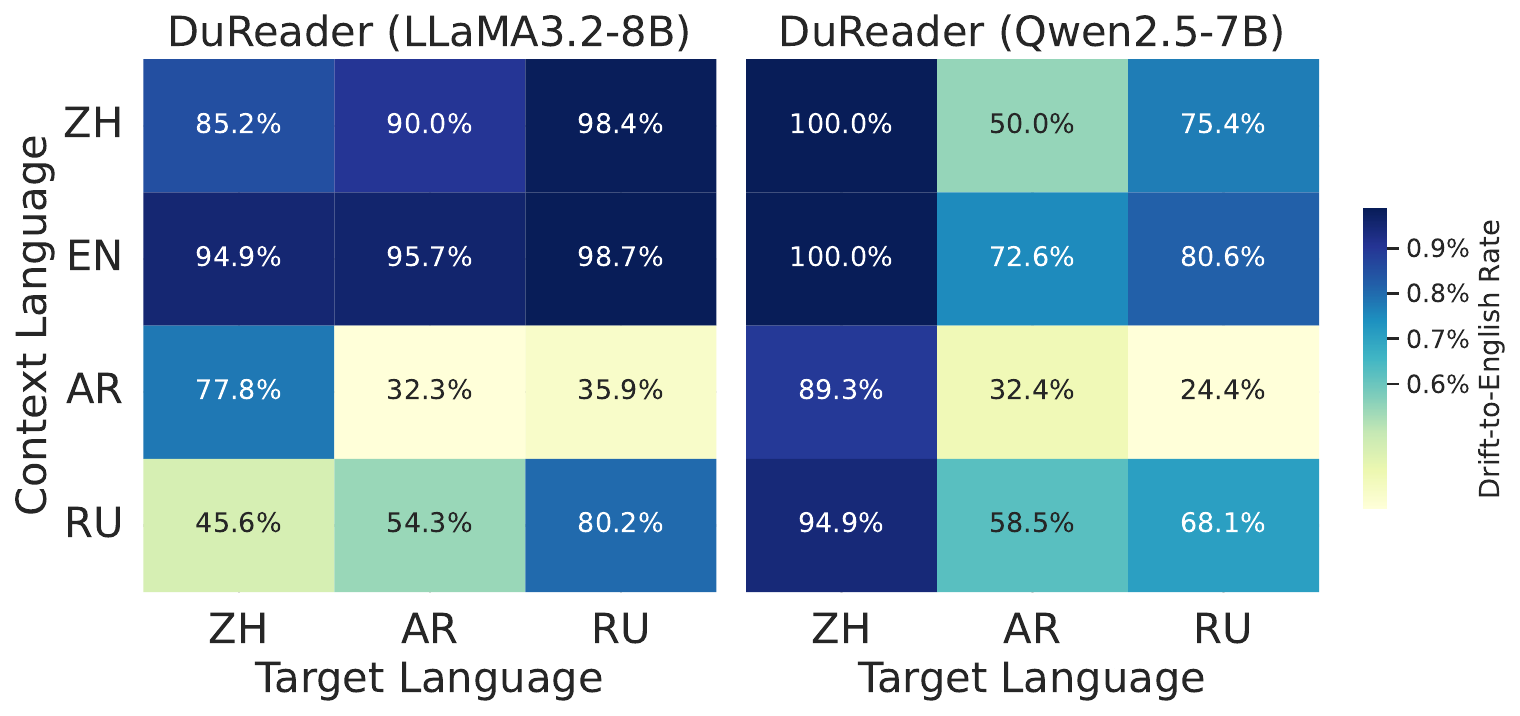}
	\caption{ Language drift patterns on the DuReader dataset for LLaMA3-8B and Qwen2.5-7B models. Each cell shows the percentage of inconsistent outputs that are generated in English. }
	\label{pic:heatmap_DurReader}
\end{figure}

To validate the generality of our fallback observations in Section~\ref{sec:fallback}, we present language drift patterns on the \textbf{Musique} and \textbf{DuReader} datasets in Figures~\ref{pic:heatmap_musique} and~\ref{pic:heatmap_DurReader}. Across both datasets, we observe consistent fallback-to-English behavior, though with greater variability compared to HotpotQA.

In \textbf{Musique}, we find that most drifted generations—across all target languages—are still predominantly in English. For example, \texttt{LLaMA3-8B} produces over 90\% English outputs even under typologically distant pairs such as \texttt{RU-–ZH} and \texttt{AR}–\texttt{RU}. \texttt{Qwen2.5-7B}, while showing slightly lower fallback rates in certain conditions (e.g., 45.0\% under \texttt{AR}-\texttt{AR}), still defaults to English in the majority of inconsistent cases across settings.

In \textbf{DuReader}, the fallback-to-English pattern remains dominant but reveals greater variability. \texttt{LLaMA3-8B} maintains high fallback rates (e.g., 98.4\% under \texttt{ZH}–\texttt{RU}, 94.9\% under \texttt{EN}-\texttt{ZH}), though notably lower values appear under Arabic contexts (e.g., 32.3\% under \texttt{AR}–\texttt{AR}). \texttt{Qwen2.5-7B} also demonstrates strong fallback tendencies (e.g., 80.6\% under \texttt{EN}–\texttt{RU}), while showing reduced English bias under certain low-resource combinations such as \texttt{AR}–\texttt{RU} (24.4\%).

Across both datasets, English remains the most frequent fallback language in drifted outputs regardless of context-target language configuration. These results reinforce the main finding from HotpotQA: \textbf{English acts as a semantic attractor during multilingual generation}, driven by token-level priors and model-internal biases.

\subsection*{Appendix B: Supplementary Drift Type Analysis on Musique and DuReader.}\label{app:heatmap}

\begin{figure}[h]
	\centering
	\includegraphics[width=0.99\linewidth]{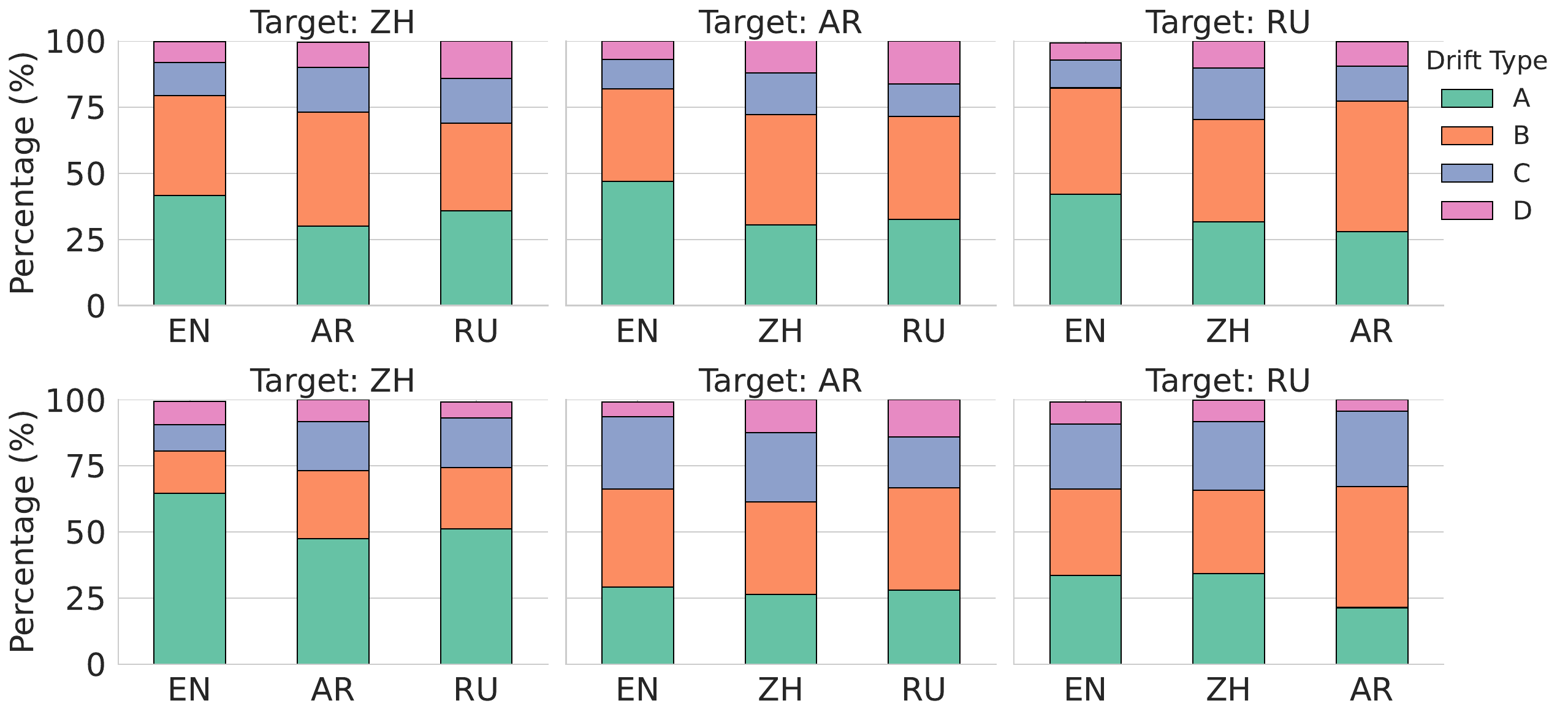}
	\caption{Distribution of four language drift types across different target–context language pairs in the Musique dataset. Each subplot corresponds to a fixed target language (ZH, AR, RU), with the x-axis denoting the context language. The top row displays results for LLaMA3-8B, and the bottom row for Qwen2.5-7B.}
	\label{pic:drift_type_musique}
\end{figure}

\begin{figure}[h]
	\centering
	\includegraphics[width=0.99\linewidth]{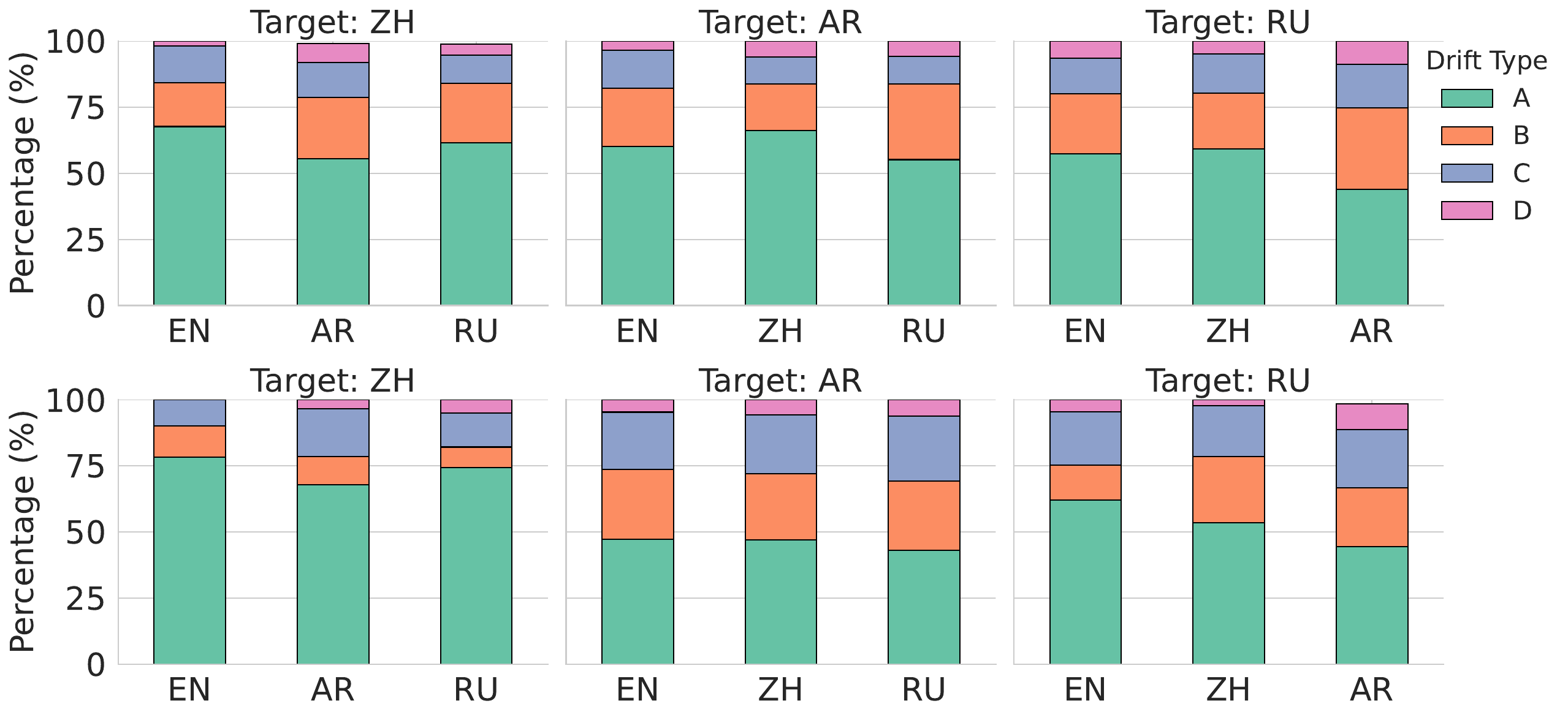}
	\caption{Distribution of four language drift types across different target–context language pairs in the DuReader dataset. Each subplot corresponds to a fixed target language (ZH, AR, RU), with the x-axis denoting the context language. The top row displays results for LLaMA3-8B, and the bottom row for Qwen2.5-7B.}
	\label{pic:drift_type_dureader}
\end{figure}

\begin{table*}[t]
\centering
\small
\begin{tabular}{c|p{3.6cm}|p{5.3cm}|p{4.8cm}}
\hline
\textbf{Type} & \textbf{Name} & \textbf{Definition} & \textbf{Impact} \\
\hline
Type A & Named Entity Representation Divergence & Inconsistent spelling or transliteration of named entities across languages. & Surface-level mismatch despite semantic equivalence. \\
\hline
Type B & Answer Target Shift (Granularity / Focus) & Answer scope or specificity shifts, or distractor-language terms mislead attention. & Plausible but incorrect answer due to focus drift. \\
\hline
Type C & Reasoning Chain Misalignment & Reasoning path is disrupted due to mixed-language context or code-switching. & Logical inconsistency, premature or broken inference. \\
\hline
Type D & Conceptual Reference Shift & Background semantics or cultural cues activate unintended knowledge concepts. & Factually incorrect but fluent outputs. \\
\hline
\end{tabular}
\caption{Taxonomy of language-induced reasoning drift types observed in multilingual Chain-of-Thought generation. Each type captures a distinct failure mode driven by multilingual input interference.}
\label{tab:drift_types}
\end{table*}

To confirm the robustness of our drift taxonomy across datasets, we analyze the distribution of four language drift types on \textbf{Musique} and \textbf{DuReader}, as shown in Figures~\ref{pic:drift_type_musique} and~\ref{pic:drift_type_dureader}. Consistent with the patterns observed in HotpotQA (Figure~\ref{pic:drift_type}), the majority of drifted outputs are dominated by \textbf{Type A: Named Entity Representation Divergence}, followed by \textbf{Type B: Answer Target Shift}, with \textbf{Type C} and \textbf{Type D} occurring less frequently.

On both datasets, \texttt{LLaMA3-8B} and \texttt{Qwen2.5-7B} exhibit similar relative proportions across all target languages (\texttt{ZH}, \texttt{AR}, \texttt{RU}), indicating the generalizability of our categorization. For example, in Musique, Type A accounts for over 50\% of drifted outputs in most configurations, particularly under typologically mismatched contexts such as \texttt{ZH}–\texttt{AR} or \texttt{AR}–\texttt{RU}. Type B remains the second most common error mode, often reflecting shifts in answer scope or misalignment introduced by distractor-language cues. Types C (reasoning chain misalignment) and D (conceptual reference shift) appear less frequently but are more prominent under challenging cross-lingual settings, such as \texttt{RU}–\texttt{ZH} or \texttt{AR}–Z\texttt{H}.

The trends in DuReader further reinforce this taxonomy. Across both models, Type A consistently accounts for the largest share of drifted outputs, with Qwen2.5-7B again showing a slightly stronger skew toward Type A, particularly in Chinese and Arabic targets. While the absolute proportions vary modestly across datasets, the relative ordering of drift types is stable.

These results confirm that language drift is not a monolithic phenomenon, but manifests in structured and interpretable patterns across different multilingual settings. The consistent dominance of surface-level divergences (Type A) and granularity shifts (Type B) suggests that decoding-time interventions should particularly target entity and answer-level alignment, while maintaining robustness against deeper semantic perturbations.

\subsection*{Appendix C: Drift Type Taxonomy.}\label{app:type}
To support our qualitative and quantitative analysis of output language drift in multilingual reasoning, we provide a detailed taxonomy of drift behaviors in Table~\ref{tab:drift_types}. Each type captures a distinct failure mode that arises from cross-lingual interference during generation, particularly in Chain-of-Thought (CoT) decoding.

\begin{itemize}
    \item \textbf{Type A: Named Entity Representation Divergence} refers to inconsistent spelling, transliteration, or surface form of named entities across languages. This results in semantically correct but surface-mismatched outputs, and is often the most frequent drift type observed.
    \item \textbf{Type B: Answer Target Shift} describes cases where the model answers a semantically adjacent but incorrect sub-target, often due to focus drift or misleading cues from distractor-language evidence. These outputs are typically plausible but factually wrong.
    \item \textbf{Type C: Reasoning Chain Misalignment} involves logical disruption in the CoT path, such as abrupt switching between languages, premature termination, or reasoning loops. These are less frequent but critical, as they undermine answer validity.
    \item \textbf{Type D: Conceptual Reference Shift} captures culturally or semantically biased generations, where background knowledge in the dominant language (e.g., English) activates unintended associations. Though rare, these often produce fluent but incorrect content.
\end{itemize}

This taxonomy enables structured annotation and interpretation of drifted outputs, and informs the design of targeted mitigation strategies. As shown in Figures~\ref{fig:drifttype-hotpot}–\ref{fig:drifttype-dureader}, the relative prevalence of these types is consistent across datasets and models, with A and B accounting for the majority of drift cases.

\subsection*{Appendix D: ICL Improves Performance but Undermines Consistency}\label{app:ICL}

Table 7 reports the full evaluation metrics across all target–context language pairs, with and without ICL, on the \textbf{HotpotQA}, \textbf{Musique}, and \textbf{DuReader} datasets. These results correspond to the high-level trends shown in Figure~\ref{pic:radar} and provide detailed quantitative insights into how ICL influences both task performance and output language consistency. Across datasets, we observe consistent trends that echo our radar plot findings:

\begin{itemize}
    \item \textbf{ICL significantly improves task accuracy} as measured by ROUGE and BLEU. For example, in HotpotQA under the \texttt{ZH–EN} condition, ROUGE improves from 0.124 to 0.182, and BLEU from 0.054 to 0.086.
    
    \item \textbf{ICL often reduces language consistency} in cross-lingual settings. Notably, in the \texttt{ZH–EN} case, LC drops from 0.883 (w/o ICL) to 0.684 (w/ ICL), despite better content metrics—confirming that improved reasoning can come at the cost of linguistic fidelity.
    
    \item \textbf{The LC drop is most severe for typologically distant language pairs}, particularly when English is the context language for non-English targets. For instance, in Musique under \texttt{AR–EN}, LC drops from 0.992 to 0.774.
    
    \item \textbf{The EN target language is highly robust} across configurations, maintaining near-perfect LC and strong ROUGE/BLEU gains even under multilingual contexts (e.g., \texttt{EN–ZH}, \texttt{EN–AR}).
\end{itemize}

\begin{figure*}[h]
    \centering
    \includegraphics[width=\textwidth]{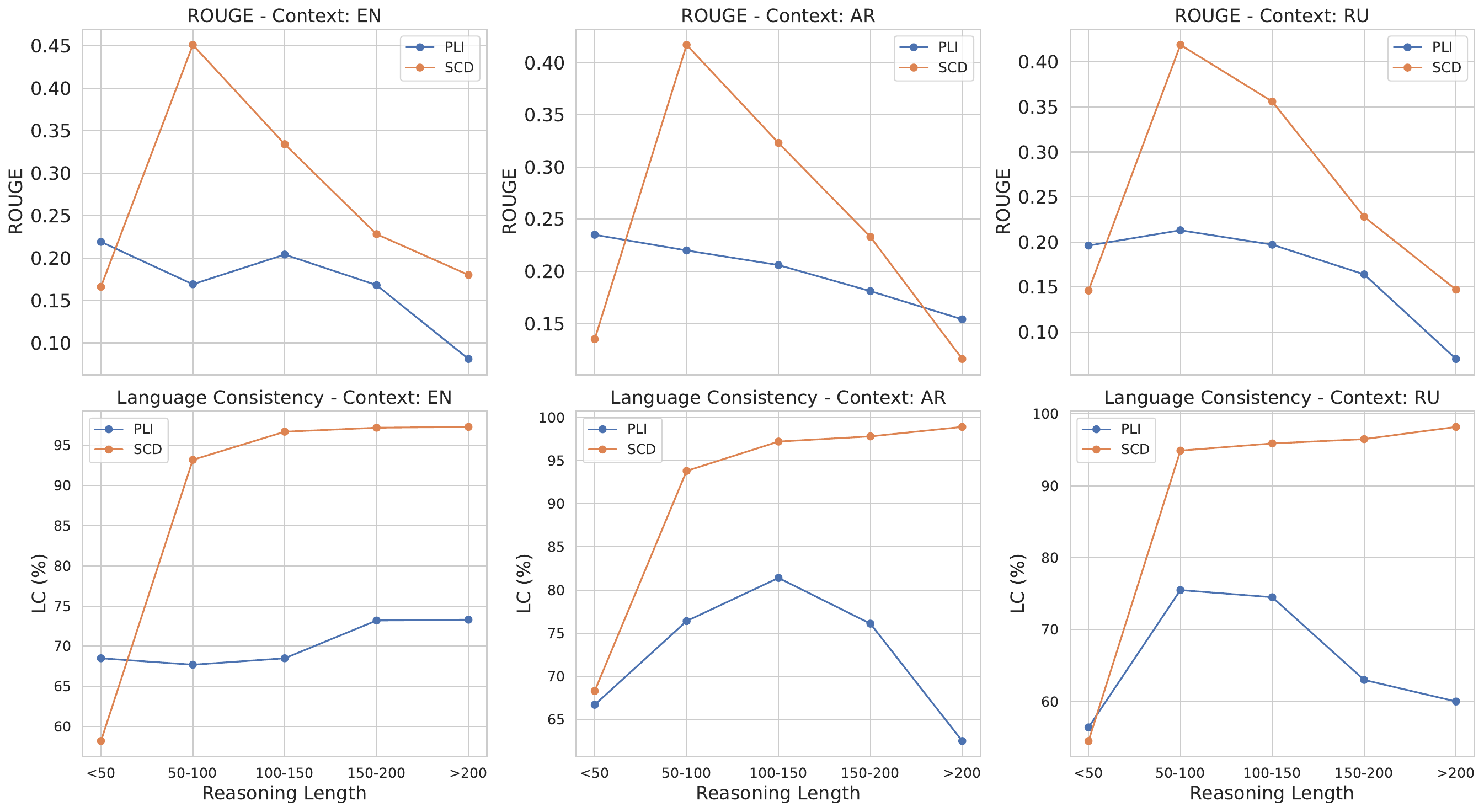}
    \caption{
    ROUGE and LC scores of \textbf{Prompted Language Instruction (PLI)} and \textbf{Soft-Constrained Decoding (SCD)} across different reasoning length intervals on the HotpotQA dataset (target language: \texttt{ZH}). Each column corresponds to a different context language (\texttt{EN}, \texttt{AR}, \texttt{RU}). SCD consistently outperforms PLI across all bins, with the most significant gains observed in the 50--150 token range. LC improvements are particularly dramatic, with SCD achieving over 95\% consistency even under strong cross-lingual interference.}
    \label{fig:length_analysis}
\end{figure*}

These complete results support the conclusion that while ICL enhances factual correctness and reasoning depth, it amplifies exposure to cross-lingual evidence, thereby increasing the risk of output language drift.

\subsection*{Appendix E: Decoding Strategy Comparison on Qwen2.5-7B.}\label{app:main_results}
Table 6 presents the full evaluation results of three decoding strategies—Prompted Language Instruction, Translation-Based Evaluation, and our proposed Soft-Constrained Decoding (SCD)—on \texttt{Qwen2.5-7B} across the HotpotQA, Musique, and DuReader datasets. These results complement Table~\ref{tab:main_results} for \texttt{LLaMA3-8B}, and confirm that the benefits of SCD generalize across different model architectures.

Consistent with our findings on \texttt{LLaMA3-8B}, we observe the following trends:

\begin{itemize}
    \item \textbf{Prompted-only decoding} leads to moderate task performance but suffers from substantial language drift. For example, on HotpotQA with \texttt{ZH–EN}, LC drops to 78.0\%, despite achieving a ROUGE score of 0.331.

    \item \textbf{Translation-based evaluation} achieves perfect LC by construction (100\% for all settings), but generally underperforms SCD in ROUGE and BLEU. This highlights that post-hoc translation does not recover reasoning fidelity lost during drift. For instance, in Musique \texttt{AR–EN}, SCD yields higher ROUGE (0.272 vs. 0.221) and BLEU (0.158 vs. 0.125), despite slightly lower LC (96.6\% vs. 100\%).

    \item \textbf{SCD consistently outperforms Prompted decoding} in both content quality and LC. Across nearly all context-target configurations, SCD raises LC by 5–15 percentage points compared to Prompted, while also producing longer and more accurate reasoning chains. Notably, in DuReader under \texttt{ZH–EN}, SCD improves LC from 94.0\% to 97.4\% and ROUGE from 0.447 to 0.462.
\end{itemize}

Together with LLaMA3-8B results, these findings demonstrate that \textbf{SCD provides a robust and general decoding-time intervention} that enhances both reasoning quality and language alignment across multilingual RAG settings—without requiring additional training or model modification.

\subsection*{Appendix F: Dataset Construction and Examples.}\label{app:data}
To construct multilingual versions of retrieval-augmented QA datasets (e.g., HotpotQA), we follow a semi-automatic pipeline that translates relevant data components using a high-quality GPT-based translation API (GPT-4o), followed by structural reorganization. Specifically, for each original English example, we translate the query, answer, supporting context into three target languages: Chinese (ZH), Russian (RU), and Arabic (AR). This results in a multilingual version of each sample, where all input components are aligned across languages. All translations are performed with instruction-based prompting to preserve semantic accuracy, named entities, and task relevance. Finally, we reorganize each translated instance into a structured multilingual format, enabling controlled experiments where the language of any input component (query, context, exemplars) can be independently varied.

\begin{figure}[h]
	\centering
	\includegraphics[width=0.99\linewidth]{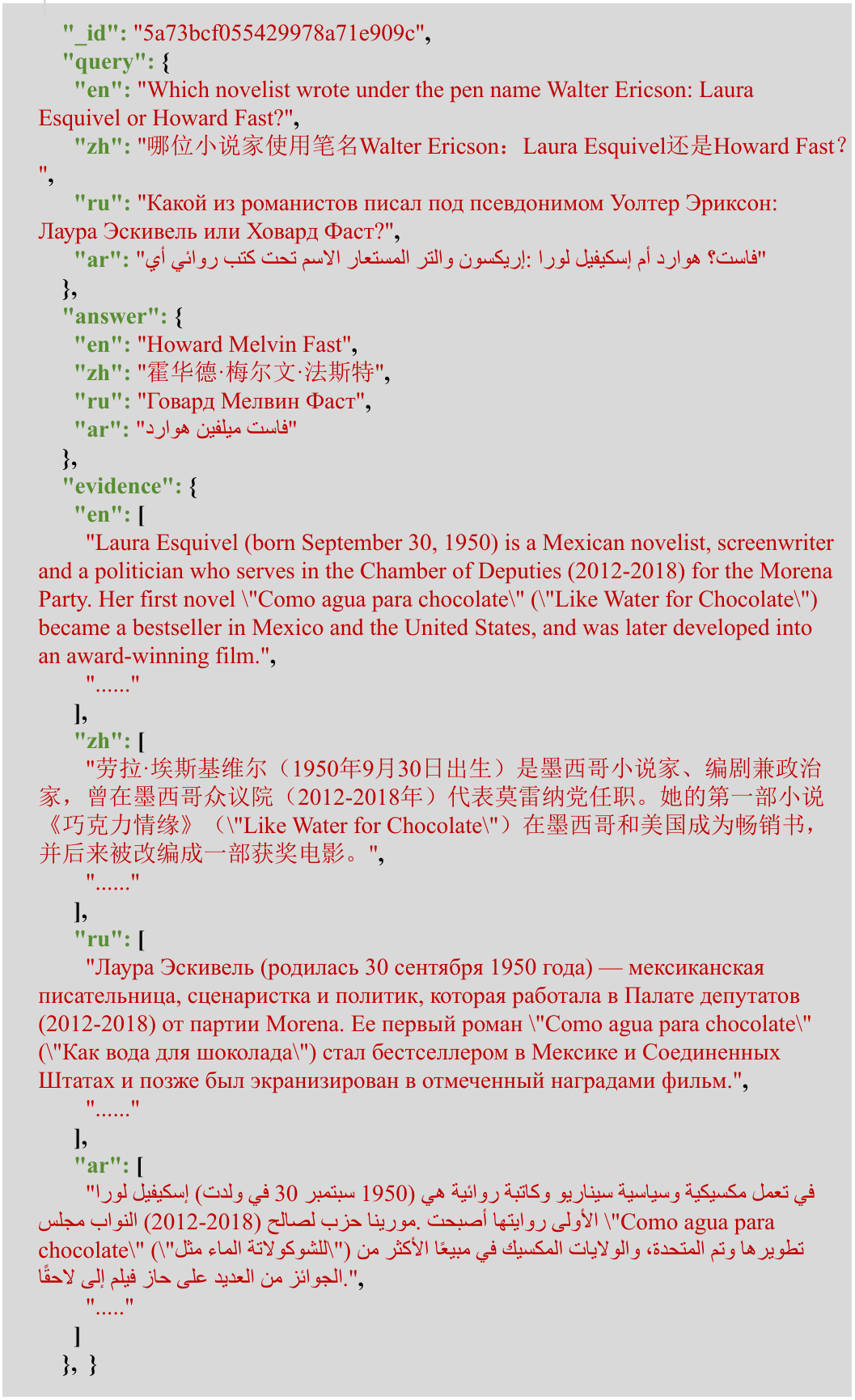}
	\caption{Example of a multilingual sample from our constructed dataset. Each instance includes queries, answers, and evidence passages in four languages: English (\texttt{en}), Chinese (\texttt{zh}), Russian (\texttt{ru}), and Arabic (\texttt{ar}). This unified JSON-style structure enables controlled evaluation and fine-grained analysis across languages.}

	\label{pic:data}
\end{figure}

\subsection*{Appendix G: Extended Analysis: Impact of Reasoning Length.}\label{app:length}

To further understand how language drift varies with the depth of reasoning, we analyze model performance across different reasoning length intervals. Specifically, we segment the generated Chain-of-Thought (CoT) outputs into five bins based on token length: $<$50, 50--100, 100--150, 150--200, and $>$200 tokens. We then compare the performance of SCD and PLI in terms of ROUGE and language consistency (LC), under three different context languages.

As shown in Figure~\ref{fig:length_analysis}, SCD consistently surpasses PLI in both ROUGE and LC across all length intervals. The performance gap is most pronounced in the mid-length range (50--150 tokens), where SCD reaches peak effectiveness. For example, in the 100--150 bin under RU context, SCD boosts LC to over 95\%, while PLI lags significantly behind. Notably, even in shorter ($<$50 tokens) and longer ($>$200 tokens) reasoning segments, where performance is generally more volatile, SCD maintains a clear advantage.

These results indicate that prompt-only language control (PLI) struggles with longer or more complex reasoning, likely due to insufficient guidance throughout the CoT trajectory. In contrast, SCD preserves target language alignment throughout the decoding process by dynamically discouraging non-target tokens, allowing the model to retain both factual correctness and linguistic fidelity. This analysis reinforces the practical value of SCD, especially for real-world multilingual applications involving long-form or multi-step reasoning. All detailed experiments are shown in Table~\ref{table:length}.

\begin{table*}[t]
\centering
\begin{tabular}{cllllllllll}
\hline
\multicolumn{1}{c|}{\textbf{}}              & \multicolumn{2}{c|}{\textbf{\textless{}50}}                          & \multicolumn{2}{c|}{\textbf{50-100}}                                 & \multicolumn{2}{c|}{\textbf{100-150}}                                & \multicolumn{2}{c|}{\textbf{150-200}}                                & \multicolumn{2}{c}{\textbf{\textgreater{}200}}                      \\ \hline
\multicolumn{1}{c|}{\textbf{Context Language}} & \multicolumn{1}{c}{\textbf{PLI}} & \multicolumn{1}{c|}{\textbf{SCD}} & \multicolumn{1}{c}{\textbf{PLI}} & \multicolumn{1}{c|}{\textbf{SCD}} & \multicolumn{1}{c}{\textbf{PLI}} & \multicolumn{1}{c|}{\textbf{SCD}} & \multicolumn{1}{c}{\textbf{PLI}} & \multicolumn{1}{c|}{\textbf{SCD}} & \multicolumn{1}{c}{\textbf{PLI}} & \multicolumn{1}{c}{\textbf{SCD}} \\ \hline
\multicolumn{11}{l}{\textit{\textbf{ROUGE}}}                                                                                                                                                                                                                                                                                                                                                                  \\ \hline
\multicolumn{1}{c|}{\textbf{EN}}            & 0.219                            & \multicolumn{1}{l|}{0.166}        & 0.169                            & \multicolumn{1}{l|}{0.451}        & 0.204                            & \multicolumn{1}{l|}{0.334}        & 0.168                            & \multicolumn{1}{l|}{0.228}        & 0.081                            & 0.180                            \\ 
\multicolumn{1}{c|}{\textbf{AR}}            & 0.235                            & \multicolumn{1}{l|}{0.135}        & 0.220                            & \multicolumn{1}{l|}{0.417}        & 0.206                            & \multicolumn{1}{l|}{0.323}        & 0.181                            & \multicolumn{1}{l|}{0.233}        & 0.154                            & 0.116                            \\ 
\multicolumn{1}{c|}{\textbf{RU}}            & 0.196                            & \multicolumn{1}{l|}{0.146}        & 0.213                            & \multicolumn{1}{l|}{0.419}        & 0.197                            & \multicolumn{1}{l|}{0.356}        & 0.164                            & \multicolumn{1}{l|}{0.228}        & 0.070                            & 0.147                            \\ \hline
\multicolumn{11}{l}{\textit{\textbf{LC}}}                                                                                                                                                                                                                                                                                                                                                                     \\ \hline
\multicolumn{1}{c|}{\textbf{EN}}            & 68.5\%                           & \multicolumn{1}{l|}{58.2\%}       & 67.7\%                           & \multicolumn{1}{l|}{93.2\%}       & 68.5\%                           & \multicolumn{1}{l|}{96.7\%}       & 73.2\%                           & \multicolumn{1}{l|}{97.2\%}       & 73.3\%                           & 97.3\%                           \\ 
\multicolumn{1}{c|}{\textbf{AR}}            & 66.7\%                           & \multicolumn{1}{l|}{68.3\%}       & 76.4\%                           & \multicolumn{1}{l|}{93.8\%}       & 81.4\%                           & \multicolumn{1}{l|}{97.2\%}       & 76.1\%                           & \multicolumn{1}{l|}{97.8\%}       & 62.5\%                           & 98.9\%                           \\ 
\multicolumn{1}{c|}{\textbf{RU}}            & 56.4\%                           & \multicolumn{1}{l|}{54.5\%}       & 75.5\%                           & \multicolumn{1}{l|}{94.9\%}       & 74.5\%                           & \multicolumn{1}{l|}{95.9\%}       & 63.0\%                           & \multicolumn{1}{l|}{96.5\%}       & 60.0\%                           & 98.2\%                           \\ \hline
\multicolumn{11}{l}{\textit{\textbf{Sample Proportion}}}                                                                                                                                                                                                                                                                                                                                                      \\ \hline
\multicolumn{1}{c|}{\textbf{EN}}            & 5.4\%                            & \multicolumn{1}{l|}{12.2\%}       & 54.2\%                           & \multicolumn{1}{l|}{22.0\%}       & 33.3\%                           & \multicolumn{1}{l|}{33.1\%}       & 5.6\%                            & \multicolumn{1}{l|}{17.8\%}       & 1.5\%                            & 14.9\%                           \\ 
\multicolumn{1}{c|}{\textbf{AR}}            & 3.6\%                            & \multicolumn{1}{l|}{10.1\%}       & 48.3\%                           & \multicolumn{1}{l|}{25.6\%}       & 37.7\%                           & \multicolumn{1}{l|}{28.9\%}       & 8.8\%                            & \multicolumn{1}{l|}{17.8\%}       & 1.6\%                            & 17.6\%                           \\ 
\multicolumn{1}{c|}{\textbf{RU}}            & 3.9\%                            & \multicolumn{1}{l|}{8.8\%}        & 47.3\%                           & \multicolumn{1}{l|}{25.7\%}       & 37.6\%                           & \multicolumn{1}{l|}{29.0\%}       & 9.2\%                            & \multicolumn{1}{l|}{20.0\%}       & 2.0\%                            & 16.5\%                           \\ \hline
\end{tabular}
\caption{
Comparison of Prompted Language Instruction (PLI) and SCD across reasoning length bins on HotpotQA, using \texttt{LLaMA3-8B-Instruct} with \texttt{ZH} as the target language. We report ROUGE, LC, and sample proportion under three context languages. SCD achieves consistent improvements across most length ranges, especially in the 50–150 token interval.
}
\end{table*}\label{table:length}

% Please add the following required packages to your document preamble:
% \usepackage{multirow}
\begin{table*}[]
\begin{tabular}{ccccccccccc}
\hline
\textbf{}                                                                               & \multicolumn{1}{c|}{\textbf{}}                                                            & \multicolumn{3}{c|}{\textbf{HotpotQA}}                            & \multicolumn{3}{c|}{\textbf{Musique}}                             & \multicolumn{3}{c}{\textbf{DuReader}}        \\ \hline
\multicolumn{1}{c|}{\textbf{\begin{tabular}[c]{@{}c@{}}Targe \\ Language\end{tabular}}} & \multicolumn{1}{c|}{\textbf{\begin{tabular}[c]{@{}c@{}}Context \\ Language\end{tabular}}} & \textbf{ROUGE} & \textbf{BLEU} & \multicolumn{1}{c|}{\textbf{LC}} & \textbf{ROUGE} & \textbf{BLEU} & \multicolumn{1}{c|}{\textbf{LC}} & \textbf{ROUGE} & \textbf{BLEU} & \textbf{LC} \\ \hline
\multicolumn{11}{l}{\textit{\textbf{Prompted Language Instruction}}}                                                                                                                                                                                                                                                                                                       \\ \hline
\multicolumn{1}{c|}{\multirow{3}{*}{\texttt{ZH}}}                                       & \multicolumn{1}{c|}{\texttt{EN}}                                                          & 0.331          & 0.196         & \multicolumn{1}{c|}{78.0\%}      & 0.317          & 0.204         & \multicolumn{1}{c|}{81.9\%}      & 0.447          & 0.270         & 94.0\%      \\
\multicolumn{1}{c|}{}                                                                   & \multicolumn{1}{c|}{\texttt{AR}}                                                          & 0.342          & 0.204         & \multicolumn{1}{c|}{86.8\%}      & 0.320          & 0.203         & \multicolumn{1}{c|}{90.3\%}      & 0.448          & 0.268         & 97.2\%      \\
\multicolumn{1}{c|}{}                                                                   & \multicolumn{1}{c|}{\texttt{RU}}                                                          & 0.337          & 0.202         & \multicolumn{1}{c|}{82.3\%}      & 0.330          & 0.214         & \multicolumn{1}{c|}{86.7\%}      & 0.445          & 0.266         & 96.1\%      \\ \hline
\multicolumn{1}{c|}{\multirow{3}{*}{\texttt{AR}}}                                       & \multicolumn{1}{c|}{\texttt{EN}}                                                          & 0.201          & 0.121         & \multicolumn{1}{c|}{86.2\%}      & 0.126          & 0.073         & \multicolumn{1}{c|}{89.0\%}      & 0.169          & 0.087         & 89.4\%      \\
\multicolumn{1}{c|}{}                                                                   & \multicolumn{1}{c|}{\texttt{ZH}}                                                          & 0.187          & 0.106         & \multicolumn{1}{c|}{89.8\%}      & 0.106          & 0.050         & \multicolumn{1}{c|}{92.0\%}      & 0.160          & 0.076         & 93.2\%      \\
\multicolumn{1}{c|}{}                                                                   & \multicolumn{1}{c|}{\texttt{RU}}                                                          & 0.206          & 0.122         & \multicolumn{1}{c|}{90.0\%}      & 0.128          & 0.070         & \multicolumn{1}{c|}{94.3\%}      & 0.174          & 0.085         & 93.5\%      \\ \hline
\multicolumn{1}{c|}{\multirow{3}{*}{\texttt{RU}}}                                       & \multicolumn{1}{c|}{\texttt{EN}}                                                          & 0.240          & 0.162         & \multicolumn{1}{c|}{83.2\%}      & 0.200          & 0.132         & \multicolumn{1}{c|}{85.7\%}      & 0.300          & 0.190         & 87.1\%      \\
\multicolumn{1}{c|}{}                                                                   & \multicolumn{1}{c|}{\texttt{ZH}}                                                          & 0.248          & 0.158         & \multicolumn{1}{c|}{88.6\%}      & 0.163          & 0.094         & \multicolumn{1}{c|}{92.7\%}      & 0.309          & 0.188         & 86.9\%      \\
\multicolumn{1}{c|}{}                                                                   & \multicolumn{1}{c|}{\texttt{AR}}                                                          & 0.248          & 0.160         & \multicolumn{1}{c|}{90.0\%}      & 0.177          & 0.107         & \multicolumn{1}{c|}{93.0\%}      & 0.323          & 0.207         & 92.8\%      \\ \hline
\multicolumn{11}{l}{\textit{\textbf{Translation-Based Evaluation}}}                                                                                                                                                                                                                                                                                                        \\ \hline
\multicolumn{1}{c|}{\multirow{3}{*}{\texttt{ZH}}}                                       & \multicolumn{1}{c|}{\texttt{EN}}                                                          & 0.352          & 0.212         & \multicolumn{1}{c|}{100.0\%}     & 0.357          & 0.229         & \multicolumn{1}{c|}{100.0\%}     & 0.455          & 0.272         & 100.0\%     \\
\multicolumn{1}{c|}{}                                                                   & \multicolumn{1}{c|}{\texttt{AR}}                                                          & 0.366          & 0.221         & \multicolumn{1}{c|}{100.0\%}     & 0.337          & 0.213         & \multicolumn{1}{c|}{100.0\%}     & 0.448          & 0.267         & 100.0\%     \\
\multicolumn{1}{c|}{}                                                                   & \multicolumn{1}{c|}{\texttt{RU}}                                                          & 0.359          & 0.215         & \multicolumn{1}{c|}{100.0\%}     & 0.351          & 0.225         & \multicolumn{1}{c|}{100.0\%}     & 0.448          & 0.266         & 100.0\%     \\ \hline
\multicolumn{1}{c|}{\multirow{3}{*}{\texttt{AR}}}                                       & \multicolumn{1}{c|}{\texttt{EN}}                                                          & 0.221          & 0.125         & \multicolumn{1}{c|}{100.0\%}     & 0.144          & 0.084         & \multicolumn{1}{c|}{100.0\%}     & 0.181          & 0.093         & 100.0\%     \\
\multicolumn{1}{c|}{}                                                                   & \multicolumn{1}{c|}{\texttt{ZH}}                                                          & 0.202          & 0.111         & \multicolumn{1}{c|}{100.0\%}     & 0.118          & 0.057         & \multicolumn{1}{c|}{100.0\%}     & 0.164          & 0.079         & 100.0\%     \\
\multicolumn{1}{c|}{}                                                                   & \multicolumn{1}{c|}{\texttt{RU}}                                                          & 0.220          & 0.122         & \multicolumn{1}{c|}{100.0\%}     & 0.134          & 0.072         & \multicolumn{1}{c|}{100.0\%}     & 0.181          & 0.090         & 100.0\%     \\ \hline
\multicolumn{1}{c|}{\multirow{3}{*}{\texttt{RU}}}                                       & \multicolumn{1}{c|}{\texttt{EN}}                                                          & 0.262          & 0.155         & \multicolumn{1}{c|}{100.0\%}     & 0.222          & 0.149         & \multicolumn{1}{c|}{100.0\%}     & 0.323          & 0.204         & 100.0\%     \\
\multicolumn{1}{c|}{}                                                                   & \multicolumn{1}{c|}{\texttt{ZH}}                                                          & 0.257          & 0.145         & \multicolumn{1}{c|}{100.0\%}     & 0.175          & 0.104         & \multicolumn{1}{c|}{100.0\%}     & 0.326          & 0.201         & 100.0\%     \\
\multicolumn{1}{c|}{}                                                                   & \multicolumn{1}{c|}{\texttt{AR}}                                                          & 0.252          & 0.144         & \multicolumn{1}{c|}{100.0\%}     & 0.184          & 0.114         & \multicolumn{1}{c|}{100.0\%}     & 0.329          & 0.210         & 100.0\%     \\ \hline
\multicolumn{11}{l}{\textit{\textbf{Soft-Constrained Decoding (Ours)}}}                                                                                                                                                                                                                                                                                                    \\ \hline
\multicolumn{1}{c|}{\multirow{3}{*}{\texttt{ZH}}}                                       & \multicolumn{1}{c|}{\texttt{EN}}                                                          & 0.376          & 0.216         & \multicolumn{1}{c|}{97.4\%}      & 0.349          & 0.221         & \multicolumn{1}{c|}{98.2\%}      & 0.462          & 0.278         & 97.9\%      \\
\multicolumn{1}{c|}{}                                                                   & \multicolumn{1}{c|}{\texttt{AR}}                                                          & 0.366          & 0.214         & \multicolumn{1}{c|}{98.2\%}      & 0.339          & 0.211         & \multicolumn{1}{c|}{98.5\%}      & 0.453          & 0.269         & 98.9\%      \\
\multicolumn{1}{c|}{}                                                                   & \multicolumn{1}{c|}{\texttt{RU}}                                                          & 0.370          & 0.215         & \multicolumn{1}{c|}{97.2\%}      & 0.349          & 0.221         & \multicolumn{1}{c|}{98.9\%}      & 0.453          & 0.268         & 98.9\%      \\ \hline
\multicolumn{1}{c|}{\multirow{3}{*}{\texttt{AR}}}                                       & \multicolumn{1}{c|}{\texttt{EN}}                                                          & 0.272          & 0.158         & \multicolumn{1}{c|}{96.6\%}      & 0.184          & 0.112         & \multicolumn{1}{c|}{99.1\%}      & 0.198          & 0.100         & 97.4\%      \\
\multicolumn{1}{c|}{}                                                                   & \multicolumn{1}{c|}{\texttt{ZH}}                                                          & 0.239          & 0.131         & \multicolumn{1}{c|}{97.1\%}      & 0.137          & 0.068         & \multicolumn{1}{c|}{98.7\%}      & 0.181          & 0.084         & 97.9\%      \\
\multicolumn{1}{c|}{}                                                                   & \multicolumn{1}{c|}{\texttt{RU}}                                                          & 0.259          & 0.147         & \multicolumn{1}{c|}{97.8\%}      & 0.173          & 0.100         & \multicolumn{1}{c|}{99.1\%}      & 0.197          & 0.096         & 98.8\%      \\ \hline
\multicolumn{1}{c|}{\multirow{3}{*}{\texttt{RU}}}                                       & \multicolumn{1}{c|}{\texttt{EN}}                                                          & 0.329          & 0.194         & \multicolumn{1}{c|}{97.2\%}      & 0.221          & 0.141         & \multicolumn{1}{c|}{98.6\%}      & 0.325          & 0.201         & 97.7\%      \\
\multicolumn{1}{c|}{}                                                                   & \multicolumn{1}{c|}{\texttt{ZH}}                                                          & 0.348          & 0.185         & \multicolumn{1}{c|}{97.3\%}      & 0.186          & 0.105         & \multicolumn{1}{c|}{98.6\%}      & 0.335          & 0.201         & 97.3\%      \\
\multicolumn{1}{c|}{}                                                                   & \multicolumn{1}{c|}{\texttt{AR}}                                                          & 0.320          & 0.181         & \multicolumn{1}{c|}{97.7\%}      & 0.193          & 0.113         & \multicolumn{1}{c|}{99.1\%}      & 0.340          & 0.214         & 97.5\%      \\ \hline
\end{tabular}
\caption{
Evaluation results of three decoding strategies—Prompted Language Instruction, Translation-Based Evaluation, and SCD on \texttt{Qwen2.5-7B} across HotpotQA, Musique, and DuReader. This table complements Table~\ref{tab:main_results} (LLaMA3-8B results), demonstrating that SCD achieves consistent improvements in both content quality and output language alignment across models. }
\end{table*}\label{tab:decoding-qwen}

\begin{table*}[t]
\centering
\begin{tabular}{ccccccccccc}
\hline
\multicolumn{1}{c|}{\textbf{Dataset}}                                                   & \multicolumn{1}{c|}{\textbf{}}                                                            & \multicolumn{3}{c|}{\textbf{HotpotQA}}                                    & \multicolumn{3}{c|}{\textbf{Musique}}                                     & \multicolumn{3}{c}{\textbf{DuReader}}                \\ \hline
\multicolumn{1}{c|}{\textbf{\begin{tabular}[c]{@{}c@{}}Targe \\ Language\end{tabular}}} & \multicolumn{1}{c|}{\textbf{\begin{tabular}[c]{@{}c@{}}Context \\ Language\end{tabular}}} & \textbf{ROUGE} & \textbf{BLEU} & \multicolumn{1}{c|}{\textbf{LC}} & \textbf{ROUGE} & \textbf{BLEU} & \multicolumn{1}{c|}{\textbf{LC}} & \textbf{ROUGE} & \textbf{BLEU} & \textbf{LC} \\ \hline
\multicolumn{8}{l}{\textit{\textbf{w/o. ICL}}}                                                                                                                                                                                                                                                                                              & \multicolumn{3}{c}{}                                 \\ \hline
\multicolumn{1}{c|}{\multirow{4}{*}{\texttt{EN}}}                                       & \multicolumn{1}{c|}{\texttt{EN}}                                                          & 0.292              & 0.142             & \multicolumn{1}{c|}{0.999}       & 0.177              & 0.084             & \multicolumn{1}{c|}{1}           & 0.201              & 0.089             & 0.969       \\
\multicolumn{1}{c|}{}                                                                   & \multicolumn{1}{c|}{\texttt{ZH}}                                                          & 0.239              & 0.114             & \multicolumn{1}{c|}{0.998}       & 0.13               & 0.053             & \multicolumn{1}{c|}{0.999}       & 0.17               & 0.064             & 0.983       \\
\multicolumn{1}{c|}{}                                                                   & \multicolumn{1}{c|}{\texttt{AR}}                                                          & 0.237              & 0.106             & \multicolumn{1}{c|}{0.999}       & 0.123              & 0.049             & \multicolumn{1}{c|}{0.999}       & 0.171              & 0.065             & 0.983       \\
\multicolumn{1}{c|}{}                                                                   & \multicolumn{1}{c|}{\texttt{RU}}                                                          & 0.237              & 0.109             & \multicolumn{1}{c|}{1}           & 0.129              & 0.052             & \multicolumn{1}{c|}{0.997}       & 0.17               & 0.065             & 0.982       \\ \hline
\multicolumn{1}{c|}{\multirow{4}{*}{\texttt{ZH}}}                                       & \multicolumn{1}{c|}{\texttt{ZH}}                                                          & 0.238              & 0.13              & \multicolumn{1}{c|}{0.981}       & 0.177              & 0.092             & \multicolumn{1}{c|}{0.979}       & 0.303              & 0.158             & 0.997       \\
\multicolumn{1}{c|}{}                                                                   & \multicolumn{1}{c|}{\texttt{EN}}                                                          & 0.124              & 0.054             & \multicolumn{1}{c|}{0.883}       & 0.091              & 0.037             & \multicolumn{1}{c|}{0.87}        & 0.186              & 0.073             & 0.96        \\
\multicolumn{1}{c|}{}                                                                   & \multicolumn{1}{c|}{\texttt{AR}}                                                          & 0.156              & 0.074             & \multicolumn{1}{c|}{0.93}        & 0.101              & 0.04              & \multicolumn{1}{c|}{0.919}       & 0.184              & 0.07              & 0.987       \\
\multicolumn{1}{c|}{}                                                                   & \multicolumn{1}{c|}{\texttt{RU}}                                                          & 0.127              & 0.059             & \multicolumn{1}{c|}{0.903}       & 0.082              & 0.032             & \multicolumn{1}{c|}{0.894}       & 0.184              & 0.07              & 0.983       \\ \hline
\multicolumn{1}{c|}{\multirow{4}{*}{\texttt{AR}}}                                       & \multicolumn{1}{c|}{\texttt{AR}}                                                          & 0.156              & 0.078             & \multicolumn{1}{c|}{0.993}       & 0.045              & 0.099             & \multicolumn{1}{c|}{0.992}       & 0.143              & 0.063             & 0.989       \\
\multicolumn{1}{c|}{}                                                                   & \multicolumn{1}{c|}{\texttt{EN}}                                                          & 0.119              & 0.055             & \multicolumn{1}{c|}{0.976}       & 0.065              & 0.023             & \multicolumn{1}{c|}{0.98}        & 0.113              & 0.043             & 0.994       \\
\multicolumn{1}{c|}{}                                                                   & \multicolumn{1}{c|}{\texttt{ZH}}                                                          & 0.113              & 0.049             & \multicolumn{1}{c|}{0.98}        & 0.059              & 0.018             & \multicolumn{1}{c|}{0.979}       & 0.11               & 0.04              & 0.977       \\
\multicolumn{1}{c|}{}                                                                   & \multicolumn{1}{c|}{\texttt{RU}}                                                          & 0.115              & 0.051             & \multicolumn{1}{c|}{0.978}       & 0.056              & 0.016             & \multicolumn{1}{c|}{0.982}       & 0.115              & 0.043             & 0.992       \\ \hline
\multicolumn{1}{c|}{\multirow{4}{*}{\texttt{RU}}}                                       & \multicolumn{1}{c|}{\texttt{RU}}                                                          & 0.193              & 0.106             & \multicolumn{1}{c|}{0.991}       & 0.108              & 0.054             & \multicolumn{1}{c|}{0.985}       & 0.193              & 0.095             & 0.995       \\
\multicolumn{1}{c|}{}                                                                   & \multicolumn{1}{c|}{\texttt{EN}}                                                          & 0.142              & 0.072             & \multicolumn{1}{c|}{0.969}       & 0.079              & 0.032             & \multicolumn{1}{c|}{0.966}       & 0.167              & 0.072             & 0.968       \\
\multicolumn{1}{c|}{}                                                                   & \multicolumn{1}{c|}{\texttt{ZH}}                                                          & 0.154              & 0.076             & \multicolumn{1}{c|}{0.977}       & 0.088              & 0.036             & \multicolumn{1}{c|}{0.985}       & 0.166              & 0.07              & 0.991       \\
\multicolumn{1}{c|}{}                                                                   & \multicolumn{1}{c|}{\texttt{AR}}                                                          & 0.168              & 0.087             & \multicolumn{1}{c|}{0.98}        & 0.083              & 0.033             & \multicolumn{1}{c|}{0.984}       & 0.171              & 0.072             & 0.998       \\ \hline
\multicolumn{11}{l}{\textit{\textbf{with. ICL}}}                                                                                                                                                                                                                                                                                                                                                   \\ \hline
\multicolumn{1}{c|}{\multirow{4}{*}{\texttt{EN}}}                                       & \multicolumn{1}{c|}{\texttt{EN}}                                                          & 0.601              & 0.317             & \multicolumn{1}{c|}{0.999}       & 0.504              & 0.334             & \multicolumn{1}{c|}{0.999}       & 0.385              & 0.206             & 0.985       \\
\multicolumn{1}{c|}{}                                                                   & \multicolumn{1}{c|}{\texttt{ZH}}                                                          & 0.552              & 0.277             & \multicolumn{1}{c|}{0.997}       & 0.372              & 0.199             & \multicolumn{1}{c|}{0.999}       & 0.352              & 0.166             & 0.967       \\
\multicolumn{1}{c|}{}                                                                   & \multicolumn{1}{c|}{\texttt{AR}}                                                          & 0.553              & 0.282             & \multicolumn{1}{c|}{0.999}       & 0.356              & 0.199             & \multicolumn{1}{c|}{0.997}       & 0.349              & 0.159             & 0.982       \\
\multicolumn{1}{c|}{}                                                                   & \multicolumn{1}{c|}{\texttt{RU}}                                                          & 0.545              & 0.279             & \multicolumn{1}{c|}{0.996}       & 0.382              & 0.217             & \multicolumn{1}{c|}{0.994}       & 0.343              & 0.164             & 0.963       \\ \hline
\multicolumn{1}{c|}{\multirow{4}{*}{\texttt{ZH}}}                                       & \multicolumn{1}{c|}{\texttt{ZH}}                                                          & 0.341              & 0.212             & \multicolumn{1}{c|}{0.92}        & 0.341              & 0.23              & \multicolumn{1}{c|}{0.942}       & 0.608              & 0.4               & 0.973       \\
\multicolumn{1}{c|}{}                                                                   & \multicolumn{1}{c|}{\texttt{EN}}                                                          & 0.182              & 0.086             & \multicolumn{1}{c|}{0.684}       & 0.187              & 0.097             & \multicolumn{1}{c|}{0.639}       & 0.339              & 0.166             & 0.842       \\
\multicolumn{1}{c|}{}                                                                   & \multicolumn{1}{c|}{\texttt{AR}}                                                          & 0.211              & 0.106             & \multicolumn{1}{c|}{0.777}       & 0.181              & 0.089             & \multicolumn{1}{c|}{0.765}       & 0.358              & 0.175             & 0.901       \\
\multicolumn{1}{c|}{}                                                                   & \multicolumn{1}{c|}{\texttt{RU}}                                                          & 0.209              & 0.107             & \multicolumn{1}{c|}{0.795}       & 0.169              & 0.087             & \multicolumn{1}{c|}{0.645}       & 0.343              & 0.168             & 0.831       \\ \hline
\multicolumn{1}{c|}{\multirow{4}{*}{\texttt{AR}}}                                       & \multicolumn{1}{c|}{\texttt{AR}}                                                          & 0.366              & 0.221             & \multicolumn{1}{c|}{0.954}       & 0.218              & 0.138             & \multicolumn{1}{c|}{0.966}       & 0.284              & 0.153             & 0.97        \\
\multicolumn{1}{c|}{}                                                                   & \multicolumn{1}{c|}{\texttt{EN}}                                                          & 0.294              & 0.162             & \multicolumn{1}{c|}{0.854}       & 0.144              & 0.08              & \multicolumn{1}{c|}{0.9}         & 0.209              & 0.099             & 0.882       \\
\multicolumn{1}{c|}{}                                                                   & \multicolumn{1}{c|}{\texttt{ZH}}                                                          & 0.265              & 0.143             & \multicolumn{1}{c|}{0.884}       & 0.12               & 0.057             & \multicolumn{1}{c|}{0.892}       & 0.193              & 0.08              & 0.87        \\
\multicolumn{1}{c|}{}                                                                   & \multicolumn{1}{c|}{\texttt{RU}}                                                          & 0.28               & 0.151             & \multicolumn{1}{c|}{0.886}       & 0.121              & 0.061             & \multicolumn{1}{c|}{0.898}       & 0.186              & 0.077             & 0.895       \\ \hline
\multicolumn{1}{c|}{\multirow{4}{*}{\texttt{RU}}}                                       & \multicolumn{1}{c|}{\texttt{RU}}                                                          & 0.373              & 0.215             & \multicolumn{1}{c|}{0.895}       & 0.29               & 0.185             & \multicolumn{1}{c|}{0.938}       & 0.374              & 0.223             & 0.936       \\
\multicolumn{1}{c|}{}                                                                   & \multicolumn{1}{c|}{\texttt{EN}}                                                          & 0.333              & 0.177             & \multicolumn{1}{c|}{0.802}       & 0.218              & 0.119             & \multicolumn{1}{c|}{0.819}       & 0.285              & 0.15              & 0.843       \\
\multicolumn{1}{c|}{}                                                                   & \multicolumn{1}{c|}{\texttt{ZH}}                                                          & 0.335              & 0.172             & \multicolumn{1}{c|}{0.851}       & 0.206              & 0.102             & \multicolumn{1}{c|}{0.902}       & 0.296              & 0.149             & 0.858       \\
\multicolumn{1}{c|}{}                                                                   & \multicolumn{1}{c|}{\texttt{AR}}                                                          & 0.339              & 0.179             & \multicolumn{1}{c|}{0.868}       & 0.214              & 0.109             & \multicolumn{1}{c|}{0.925}       & 0.288              & 0.143             & 0.909       \\ \hline
\end{tabular}
\caption{
Detailed evaluation results for all target–context language pairs on HotpotQA, Musique, and DuReader using \texttt{LLaMA3-8B}, with and without ICL. Each block reports ROUGE, BLEU, and language consistency (LC) under fixed target language (rows) and varying context language (columns). These results complement the radar plot in Figure~\ref{pic:radar}, revealing that while ICL improves content metrics across the board, it often reduces language consistency in cross-lingual settings—especially when English evidence is introduced. }
\end{table*}\label{tab:icl_detailed_llama}

\end{document}